 \documentclass[final,3p,times,twocolumn]{elsarticle}
 \usepackage[utf8]{inputenc}

 \makeatletter
\def\ps@pprintTitle{%
    \let\@oddhead\@empty
    \let\@evenhead\@empty
    \def\@oddfoot{\footnotesize\itshape
         {~} \hfill\today}%
    \let\@evenfoot\@oddfoot
    }
\makeatother

\usepackage[figuresright]{rotating}
\usepackage{adjustbox}
\usepackage{setspace}
\usepackage{tocloft}
\usepackage{mathtools}
\usepackage{eurosym}
\usepackage{fancyhdr}
\usepackage{algpseudocode} 
\usepackage{textcomp}
\usepackage{tikz}
\usepackage{subcaption}
\usepackage{booktabs}
\usepackage{placeins}
\usepackage{graphicx}
\usepackage{color}
\usepackage{array}
\usepackage{multirow}
\usepackage{url} 
\usepackage{needspace}
\usepackage[printonlyused]{acronym}
\usepackage{float}
\usepackage{rotating}
\usepackage[titletoc,toc,page]{appendix}
\usepackage{lipsum}
\usepackage{todonotes}
\usepackage{algorithm}
\usepackage{algpseudocode}
\usepackage{mdframed}
\usepackage{nomencl}
\usepackage{multicol}

\makenomenclature

\usepackage{etoolbox}
\renewcommand\nomgroup[1]{%
  \item[\bfseries
  \ifstrequal{#1}{A}{Coefficients}{%
  \ifstrequal{#1}{B}{Main variables}{%
  \ifstrequal{#1}{C}{Models}{%
  \ifstrequal{#1}{D}{Other abbreviations}{}}}}%
]}

\newcommand\MyBox[2]{
  \fbox{\lower0.75cm
    \vbox to 1cm{\vfil
      \hbox to 1cm{\hfil\parbox{1.4cm}{#1\\#2}\hfil}
      \vfil}%
  }%
}
\usepackage{amssymb}

 \usepackage{lineno}

\let\today\relax
\makenomenclature

\makeatletter
\newcommand{\closenomencl}{%
  \closeout\@nomenclaturefile%
}
\makeatother

\newcommand{\writenomencl}[1]{%
  \closenomencl%
  \IfFileExists{#1.nlo}{%
    \write18{%
      makeindex -s nomencl.ist -o #1.nls -t #1.nlg #1.nlo%
    }%
  }{\typeout{Nothing there}}%
}

\AtEndDocument{\writenomencl{\jobname}}

\begin{document}

\begin{frontmatter}





\title{Opening the Black Box: Towards inherently interpretable energy data imputation models using building physics insight}


 \author[label1]{Antonio Liguori*$^{,}$} \author[label2]{Matias Quintana} 
 \author[label3]{Chun Fu}
 \author[label3]{Clayton Miller} \author[label1]{J\'{e}r\^{o}me Frisch} \author[label1]{Christoph van Treeck}   

\address[label1]{E3D - Institute of Energy Efficiency and Sustainable Building, RWTH Aachen University, Mathieustr. 30, 52074 Aachen, Germany}

\address[label2]{Future Cities Laboratory Global, Singapore-ETH Centre, Singapore}
\address[label3]{Department of the Built Environment, College of Design and Engineering, National University of Singapore (NUS), Singapore}

\begin{abstract}

Missing data are frequently observed by practitioners and researchers in the building energy modeling community. In this regard, advanced data-driven solutions, such as Deep Learning methods, are typically required to reflect the non-linear behavior of these anomalies. As an ongoing research question related to Deep Learning, a model's applicability to limited data settings can be explored by introducing prior knowledge in the network. This same strategy can also lead to more interpretable predictions, hence facilitating the field application of the approach. For that purpose, the aim of this paper is to propose the use of Physics-informed Denoising Autoencoders (PI-DAE) for missing data imputation in commercial buildings. In particular, the presented method enforces physics-inspired soft constraints to the loss function of a Denoising Autoencoder (DAE). In order to quantify the benefits of the physical component, an ablation study between different DAE configurations is conducted. First, three univariate DAEs are optimized separately on indoor air temperature, heating, and cooling data. Then, two multivariate DAEs are derived from the previous configurations. Eventually, a building thermal balance equation is coupled to the last multivariate configuration to obtain PI-DAE. Additionally, two commonly used benchmarks are employed to support the findings. It is shown how introducing physical knowledge in a multivariate Denoising Autoencoder can enhance the inherent model interpretability through the optimized physics-based coefficients. While no significant improvement is observed in terms of reconstruction error with the proposed PI-DAE, its enhanced robustness to varying rates of missing data and the valuable insights derived from the physics-based coefficients create opportunities for wider applications within building systems and the built environment.


\end{abstract}

\end{frontmatter}
\onecolumn

\FloatBarrier

\begin{table}[H]
\begin{mdframed}
\begin{multicols}{2}
\nomenclature[A]{\(a\)}{Physics-based coefficient for the outdoor air temperature}
\nomenclature[A]{\(b\)}{Physics-based coefficient for the cooling flow rate}
\nomenclature[A]{\(c\)}{Physics-based coefficient for the heating flow rate}
\nomenclature[A]{\(PCC_1\)}{Pearson Correlation Coefficient between the indoor air temperature and cooling flow rate}
\nomenclature[A]{\(PCC_2\)}{Pearson Correlation Coefficient between the indoor air temperature and the heating flow rate}
\nomenclature[A]{\(PCC_3\)}{Pearson Correlation Coefficient between the heating flow rate and cooling flow rate}
\nomenclature[A]{\(PCC_4\)}{Pearson Correlation Coefficient between the correlation coefficient between the outdoor air temperature and cooling flow rate}
\nomenclature[A]{\(PCC_5\)}{Pearson Correlation Coefficient between the correlation coefficient between the outdoor air temperature and heating flow rate}
\nomenclature[A]{\(PCC_6\)}{Pearson Correlation Coefficient between the correlation coefficient between the outdoor air temperature and indoor air temperature}
\nomenclature[B]{\(T_{sa}\)}{Roof terminal units supply air temperature}
\nomenclature[B]{\(T_{ra}\)}{Roof terminal units return air temperature}
\nomenclature[B]{\(T_{ma}\)}{Roof terminal units mixed air temperature}
\nomenclature[B]{\(T_{oa}\)}{Roof terminal units outdoor air temperature}
\nomenclature[B]{\(\dot{V}_{sa}\)}{Roof terminal units filtered supply air volume flow rate}
\nomenclature[B]{\(T_{shw}\)}{Heat pump supply hot water temperature}
\nomenclature[B]{\(T_{rhw}\)}{Heat pump return hot water temperature}
\nomenclature[B]{\(\dot{V}_{shw}\)}{Heat pump supply hot water volume flow rate}
\nomenclature[B]{\(\dot{Q}_{cool_{tot}}\)}{Total cooling flow rate}
\nomenclature[B]{\(\dot{Q}_{hw}\)}{Reheat water flow rate at the heat pump}
\nomenclature[B]{\(T_{ra_{avg}}\)}{Average indoor air temperature}
\nomenclature[B]{\(T_{oa_{avg}}\)}{Average outdoor air temperature}
\nomenclature[C]{\(Univariate\_DAE\_1\)}{Univariate Denoising Autoencoder for the indoor air temperature}
\nomenclature[C]{\(Univariate\_DAE\_2\)}{Univariate Denoising Autoencoder for the heating flow rate}
\nomenclature[C]{\(Univariate\_DAE\_3\)}{Univariate Denoising Autoencoder for the cooling flow rate}
\nomenclature[C]{\(Multivariate\_DAE\_1\)}{Multivariate Denoising Autoencoder for the indoor air temperature and the heating and cooling flow rates}
\nomenclature[C]{\(Multivariate\_DAE\_2\)}{Multivariate Denoising Autoencoder for the indoor and outdoor air temperature and the heating and cooling flow rates}
\nomenclature[C]{\(PI-DAE\)}{Physics-informed Denoising Autoencoder}
\nomenclature[C]{\(LIN\)}{Linear Interpolation}
\nomenclature[C]{\(KNN\)}{k-Nearest Neighbors Impute}
\nomenclature[D]{\(BEMs\)}{building energy simulation models}
\nomenclature[D]{\(CAE\)}{Convolutional Denoising Autoencoder}
\nomenclature[D]{\(Aug\)}{augmentation rate}
\nomenclature[D]{\(BEMs\)}{building energy simulation models}
\nomenclature[D]{\(CAE\)}{Convolutional Denoising Autoencoder}
\nomenclature[D]{\(CFD\)}{computational fluid dynamics}
\nomenclature[D]{\(CR\)}{corruption rate}
\nomenclature[D]{\(DAEs\)}{Denoising Autoencoders}
\nomenclature[D]{\(EGD\)}{European Green Deal}
\nomenclature[D]{\(HDDPC\)}{hierarchical data-driven predictive control}
\nomenclature[D]{\(HP\)}{heat pump}
\nomenclature[D]{\(IEQ\)}{indoor environmental quality}
\nomenclature[D]{\(IQR\)}{interquartile range}
\nomenclature[D]{\(LIME\)}{Local Interpretable Model-Agnostic Explanations}
\nomenclature[D]{\(LSTM\)}{Long Short-Term Memory}
\nomenclature[D]{\(MPC\)}{model predictive control}
\nomenclature[D]{\(ODE\)}{ordinary differential equation}
\nomenclature[D]{\(PCC\)}{Pearson Correlation Coefficient}
\nomenclature[D]{\(PCDL\)}{Physically Consistent Deep Learning }
\nomenclature[D]{\(PCNN\)}{Physically Consistent Neural Network}
\nomenclature[D]{\(PDEs\)}{partial differential equations}
\nomenclature[D]{\(PINNs\)}{Physics-informed Neural Networks}
\nomenclature[D]{\(RC\)}{resistance-capacitor}
\nomenclature[D]{\(RMSE\)}{root mean squared error}
\nomenclature[D]{\(RNN\)}{Recurrent Neural Network}
\nomenclature[D]{\(RTUs\)}{roof terminal units}
\nomenclature[D]{\(TR\)}{training rate}
\nomenclature[D]{\(UTFs\)}{underfloor terminal units}

\printnomenclature
\end{multicols}
\end{mdframed}
\end{table}
\section{Introduction}

On December 2019, the European Green Deal (EGD) was presented to the European Commission. The ultimate objective of the EGD is to achieve carbon neutrality by 2050 and to decouple economic growth from resource use \cite{fetting2020european}. As observed in the same proposal, buildings account for a major portion of the total energy consumption and greenhouse gas emissions. For this reason, a large part of the effort must go through a renovation of the old building stock. 

Building retrofit refers to the process of renovating a building during its operation phase, particularly for energy efficiency and insulation \cite{thrampoulidis2023approximating}. As such, it is of major importance in order to reduce the environmental footprint of the built environment. In the last years, the increasing number of installed meters has enabled the use of high-resolution data for building energy simulation models (BEMs) \cite{chong2021calibrating}. These data are usually used for calibration purposes in order to correctly simulate the behavior of a building \cite{angelotti2018building}. When using BEMs for retrofit analysis, the objective should be to improve both the energy efficiency and occupant thermal comfort \cite{magnier2010multiobjective}. Therefore, calibrating these models against high-resolution heating, cooling, and indoor air temperature data is an important issue \cite{ji2015bottom,claridge2006missing}.

As observed by Baltazar and Claridge \cite{baltazar2002restoration}, retrofit analysis might be hindered by the quality of the collected data. Building monitoring datasets are indeed characterized by a large number of anomalies and missing data that might be caused by a variety of reasons \cite{liguori2021indoor}. A study performed on a database containing energy usage recordings from over 600 buildings in the US concluded that around 60\% of missing data were below six consecutive hours \cite{baltazar2002restoration}. Considering that daily totals might be useful for building savings determination, incorrect estimation of these faulted data points might lead to wrong decision-making \cite{claridge2006missing}. For instance, the presence of these short-term gaps might lead to the discarding of the whole day of observation, with subsequent loss of relevant information.

\subsection{Building data imputation}

In order to avoid this scenario, imputation should be performed. Imputation is the process of replacing missing data with new estimated values \cite{donders2006gentle}. In literature, several statistical and machine learning-based imputation approaches have been proposed \cite{emmanuel2021survey}. As observed by Emmanuel et al. \cite{emmanuel2021survey}, simple statistical imputation techniques, such as the mean or linear interpolation, are often used due to their simplicity. However, these same techniques are also prone to produce inaccurate estimations. More advanced data-driven solutions, such as regression or machine learning-based imputation models, are usually necessary to achieve more stable and precise missing value replacements. 

Regression-based methods typically involve the use of a Back-propagation Neural Network trained to capture the relationships among the existing and missing features \cite{liu2015fault,li2020missing,wang2021fault}. However, the computational intensity of Back-propagation Neural Networks increases considerably when more than one variable is missing. In that case, the Expectation-Maximization algorithm has proven to be a less complex and equally effective alternative \cite{wang2021fault}. Other simplified regression methods include linear regression, K-nearest Neighbors (KNN), and Support Vector Machine \cite{fu2024filling}. When data exhibit strong autoregressive correlations, researchers tend to reformulate the data imputation problem as a prediction task. In this regard, models designed to handle sequential data, such as Recurrent Neural Networks (RNN), are typically used \cite{ahn2022comparison, hussain2022novel}. However, employing predictive methods to replace missing data limits the capability of the model by relying solely on past values \cite{festag2022generative}. In this sense, bi-directional training and encoder-decoder architectures have been shown to be reliable strategies to boost performance in imputation tasks \cite{ma2020bi,ahn2022comparison}. Nonetheless, these techniques require additional evaluations and might further complicate the model complexity and training.

Whether approaching imputation from a regression or prediction perspective, advanced data-driven methods typically pose two main challenges. First, they require an extensive amount of historical data for accurate predictions \cite{emmanuel2021survey}. In addition, they are not always readily interpretable by humans \cite{arjunan2020energystar++}. All these factors might prevent the field application of state-of-the-art imputation models by building practitioners.

\subsection{Combining physics-informed learning with building data imputation}
In the case of building energy systems, there is a variety of physical laws that pure data-driven models ignore at the beginning of the learning process \cite{yao2013state}. Encoding such knowledge from scratch results in a significant amount of data required by the algorithm. To solve this issue, the use of machine learning for solving partial differential equations (PDEs) has been recently proposed as a valid approach \cite{karniadakis2021physics}. By encoding prior physical knowledge in the form of PDEs, these models can indeed achieve more accurate predictions by using a smaller amount of monitoring data \cite{raissi2019physics}. Physics-informed Neural Networks (PINNs), recently proposed by Raissi et al. \cite{raissi2019physics}, are a family of neural networks that embeds physical principles by applying soft constraints to the loss function. Here, the authors took advantage of the automatic differentiation principle of neural networks in order to approximate the PDEs’ residuals of generic physical problems. The results proved the effectiveness of the method when applied to different forward and inverse problems in fluid and quantum mechanics and reaction-diffusion systems \cite{raissi2019physics}. 

So far, PINNs have been applied to different fields, such as biophysics, quantum chemistry, material science, and geophysics \cite{karniadakis2021physics}. In this regard, the remarkable results observed in the literature have recently inspired the use of these models in different building applications, including control-oriented thermal dynamics modeling  \cite{gokhale2022physics,di2022physically,di2023towards,xiao2023building}, indoor environmental quality (IEQ) data prediction \cite{nagarathinam2022pacman,wang2023physics} and demand response control \cite{chen2023physics}. However, to the best of the authors' knowledge, the application to missing data imputation problems has not yet been investigated.

\subsection{Bridging the gap between academic research and industry through inherent physical interpretability}

The importance of interpretable data-driven modeling is a topic that has been long discussed in the medical community \cite{arjunan2020energystar++}. On one hand, deploying accurate machine learning models might lead to enormous economic benefits without reducing the quality of the service provided. For instance, some human vital signs, such as blood pressure or body temperature, could be used as predictors to quantify the general health status of a person. This would significantly reduce the costs of hospitalization \cite{caruana2015intelligible}. On the other hand, some ethical concerns arise, especially when the model is operating in a ``black box" manner, and the correlation between the predictors and target variable can not be understood. In this regard, the risk is to learn ``dangerous" patterns in the training data that do not follow any medical logic and that are not detectable by doctors \cite{caruana2015intelligible}. For this reason, despite being less accurate, tree-based models are usually considered more reliable approaches \cite{bishop2006pattern}.

Unlike healthcare applications, model interpretability has been neglected by building researchers for many years \cite{fan2019novel}. It has been argued that the absence of physical rules governing the available data-driven models in the literature has been one of the reasons for their scarce applicability in building retrofit \cite{deb2021review}. As explained by Fan et al. \cite{fan2019novel}, this is likely because building practitioners are mainly interested in the motivations that drive the model to make a certain prediction rather than its accuracy. When looking at the literature related to interpretable building energy management, studies associated with imputation methods are, indeed, almost scarce or non-existent. Most of the papers focus, instead, on load or power prediction \cite{chen2023interpretable}. However, the same consideration does not apply to other fields, such as medicine \cite{chen2023missing}. As a matter of fact, data-driven models are, by definition, highly dependent on the quality of the used dataset \cite{liguori2021indoor}. Consider a real-world case scenario in which an unsafe sign conversion occurs during data acquisition. For instance, a negative temperature reading might be erroneously converted into a large unsigned integer \cite{brumley2006towards}. This scenario is different from other common sources of missing values, such as sensor malfunctioning or power outages \cite{chong2016imputation}. To this end, an interpretable imputation model would not only reconstruct the errors but also provide an explanation for the imputation process in the different cases \cite{chen2023interpretable}. This capability offers additional insights into the various dynamics that generate an anomaly. As a result, it can help on-site operators assess the reasonableness of the estimated values and quantify the effects on predictive modeling \cite{chen2023missing}.

While model interpretability can be enhanced by model-agnostic approaches such as Local Interpretable Model-Agnostic Explanations (LIME) \cite{mishra2017local}, these techniques provide post-hoc data-driven explanations that may even be harmful to the decision-making process \cite{sudjianto2021designing}. In this sense, introducing physical knowledge in the deep learning-based imputation model would ensure the needed inherent transparency and modularity, yet without renouncing the accuracy of neural networks \cite{karniadakis2021physics}. It is worth noting that other methods might also be used for the same purpose. For instance, the attention mechanism has been recently used in the scope of fault detection and diagnosis for building energy management \cite{li2019novel}. However, this technique requires a nontrivial modification of the model architecture. In the scope of this paper, it is proven how physics can be introduced in a modular fashion. This implies that the decoupled model is not significantly affected either in terms of original configuration or in terms of computational complexity.

\subsection{Contribution}

When implementing an imputation model, it is particularly important that the reconstruction error is independent of the particular missing component. This is crucial to ensure the robustness of the imputation process \cite{liguori2023augmenting}. In literature, this goal has already been achieved by defining a specific data augmentation technique for Denoising Autoencoders (DAEs) \cite{liguori2023augmenting}. The presented solution consisted of stacking together synthetic copies of the same training sample with pseudo-random masking noise. The scope of this paper aims to improve the aforementioned approach by encoding prior physical knowledge. In particular, the prior physical knowledge refers to a simple resistance-capacitor (RC) model able to bind together the imputed heating, cooling, and indoor air temperature data.

To summarize, the contribution of this paper is to establish the foundations for a deep learning model for building energy data imputation that is inherently interpretable in the sense that its coefficients provide indications of whether the imputed missing values are physically consistent. More in detail, the aforementioned approach identifies multi-dimensional physically consistent correlations using the learned physics-based coefficients. This is particularly valuable for improved retrofit analysis and might facilitate the field application of advanced data-driven imputation models. The used dataset was collected in an office building located in Berkeley, California, and is available open-source \cite{luo2022three}. The major research implications and findings of this work are provided in the following points. (1) It is shown how to combine the potential of DAEs with mathematical models based on physical laws as differential equations in a modular fashion. (2) It is proved that the inherent interpretability of Convolutional Denoising Autoencoders with data augmentation (CAE + Aug), as proposed in Liguori et al. \cite{liguori2023augmenting}, can be enhanced by applying physics-inspired soft constrained to the loss function. (3) It is proved that the performance of the original ``decoupled" method is not significantly affected either in terms of imputation error or in terms of model complexity. 

The remainder of this work is organized as follows: Section \ref{sec: literature} introduces the available studies in the literature that treat the topic of physics-informed learning for building applications. Section \ref{sec: method} provides details about the used model architecture and optimization. The performed experiments are presented in Section \ref{sec:results}, and the relevant results are discussed in Section \ref{sec: discussion}. The summary and final remarks are provided in Section \ref{sec: conclusion}.



     			


\section{Literature review}
\label{sec: literature}

As discussed in the previous Section, pure data-driven models, such as standard neural networks, tend to rely heavily on the quality and size of the used data. This is generally due to the lack of initial knowledge regarding the underlying system. Physics-informed Machine Learning has been widely explored in different engineering areas in which the general system principles can be described using physical models, yet the volume of data that needs to be processed exceeds the limit of conventional data analytic methods. Among others, Physics-informed Machine Learning has been widely explored in computational fluid dynamics (CFD) \cite{wang2017physics,wu2018physics}, applied CFD in the context of climate modeling \cite{howland2019wind}, geoscience modeling \cite{karimpouli2020physics} as well as non-linear dynamic systems \cite{raissi2017physics} and heat transfer \cite{zobeiry2021physics}.

The Physics-informed Data-Driven modeling has recently gained significant attention in the building energy modeling community. In one of the pioneering works on this topic, Drgona et al. \cite{drgona2020physics} proposed a constrained Recurrent Neural Network to model buildings' thermal dynamics. The encoded underlying graph structure is inspired by the physical building components. The results pointed out that the model can be generalized well to different buildings and could represent the buildings’ thermal dynamics, including the indoor climate conditions, with high validity. 

The initial results from Drgona et al. \cite{drgona2020physics} demonstrated that applying physics-inspired soft constraints to the loss function of a neural network is beneficial for modeling the thermal dynamics of a building with limited data samples. This encouraged further research in the literature \cite{gokhale2022physics,di2022physically,di2023towards,nagarathinam2022pacman,xiao2023building,chen2023physics,wang2023physics}. Specifically, concerning control-oriented thermal dynamics modeling, Gokhale et al. \cite{gokhale2022physics} implemented a Physics-informed Neural Network to model the temporal evolution of room temperatures, power consumptions, and temperatures of the building thermal mass. In particular, two different variants were proposed. The first configuration consisted of an encoder and dynamics modules, while the second was a simple Fully Connected network. To introduce physics into the network, a grey box approach with two resistances and two capacitors (2R2C) model was used. The results showed that both network architectures were able to accurately predict the room temperature. Additionally, the models proved to be data-efficient, hence requiring less training data. Di Natale et al. \cite{di2022physically} proposed a Physically Consistent Neural Network (PCNN) for single-zone thermal modeling. The model consisted of two parts, namely, a black box and a physics-inspired module. In particular, the thermal dynamics of the building were described by means of an ordinary differential equation (ODE), which represented a simple RC model. Even if the PCNN showed similar performance to the single neural network, the results proved that the temperature prediction could remain physically consistent over the input data. The same authors extended their work to multi-zone modeling in order to consider the entire building \cite{di2023towards}. The improved PCNN could outperform the used grey-box and black-box baselines. According to the authors, unlike PINNs, PCNNs can indeed provide guarantees of physical consistency. Finally, Xiao et al. \cite{xiao2023building} presented a Physically Consistent Deep Learning (PCDL) model. Furthermore, they integrated the model into a model predictive control (MPC) controller in order to evaluate the reduction of energy consumption and improvement of thermal comfort. By providing physical guarantees, the PCDL-based controller could reduce energy consumption by a maximum of 8.9\% and improve thermal comfort by a maximum of 64\%.

In the related building control literature, PINNs have also been used for IEQ measurement prediction \cite{nagarathinam2022pacman,wang2023physics}. Specifically, Nagarathinam et al. \cite{nagarathinam2022pacman} implemented a PINN-based thermal model to predict air temperature and humidity in a building. Additionally, they validated their model for MPC. The results confirmed that the proposed approach could respect the underlying physics of the system over different conditions. Finally, the method could perform better than the Long Short-Term Memory Network (LSTM) used as a baseline and consume 24\% less energy with an increased user comfort of 2\%. Additionally, Wang and Dong \cite{wang2023physics} developed a Physics-informed Input Convex Neural Network. The model was further integrated into a hierarchical data-driven predictive control (HDDPC) for space cooling and airside coil load minimization. It was observed that the proposed network could provide physically consistent predictions of indoor air temperature and CO$_2$ data. Finally, the HDDPC could achieve a reduction of cooling and airside coil energy of about 35\% and 70\%, respectively. 

Finally, Chen et al. \cite{chen2023physics} proposed a PINN for demand response control in grid-integrated buildings. In particular, the model consisted of a Fully Connected Network with a 2R2C model to provide prior physical knowledge. This method could outperform purely data-driven models for predicting the indoor air temperature and thermal load demand of different types of buildings.

In summary, in the building energy modeling community, Physics-informed Data-Driven modeling has been mainly used for predicting the indoor environmental conditions and power consumption of a building. This has led to the first attempts to integrate these models into the MPC controllers of buildings. However, no study has been found to deal with missing data imputation problems. 

\section{Methodology}
\label{sec: method}

\subsection{Dataset description}

The used dataset was collected in a multi-story office building located in Berkeley, California \cite{luo2022three}. The building is served by four roof terminal units (RTUs) with direct expansion coils that provide cooled air to the interior and exterior spaces. Specifically, the exterior spaces are zones with external walls. Here, the cooled air is eventually reheated by means of underfloor terminal units (UTFs) supported by a heat pump (HP) and variable frequency drive pumps. For detailed information about the building and used sensors, the reader is referred to the original paper from Luo et al. \cite{luo2022three}.

Out of three years of measurements, only the last one is considered. This is based on the limited data availability concerning the heat pump operation. In particular, the selected HVAC operational variables are shown in the upper part of Table \ref{tab:hvac}. 

\begin{table}[ht]
\centering
\caption{HVAC operational data (top) and final aggregated temperature, heating and cooling load data (bottom). The maximum and minimum values and the temporal resolution of the original dataset are also represented. Data are converted from the original US customary units.}
\label{tab:hvac}

\begin{tabular}{llllll}\\
      \toprule
  \multicolumn{1}{l}{Variable} & \multicolumn{1}{l}{Description} &
  \multicolumn{1}{l}{Unit} & \multicolumn{1}{l}{Max. value}  & \multicolumn{1}{l}{Min. value}   & \multicolumn{1}{l}{Time resolution [min]}\\
\toprule
    \multicolumn{1}{l}{$T_{sa}$} &
  \multicolumn{1}{l}{RTU supply air temperature} &
  \multicolumn{1}{l}{$^\circ \text{C}$}&
  \multicolumn{1}{r}{37.11}&
  \multicolumn{1}{r}{-17.77}&
  \multicolumn{1}{r}{1.00}\\
    \multicolumn{1}{l}{$T_{ra}$} &
  \multicolumn{1}{l}{RTU return air temperature} &
  \multicolumn{1}{l}{$^\circ \text{C}$}&
  \multicolumn{1}{r}{43.05}&
  \multicolumn{1}{r}{-17.77}&
  \multicolumn{1}{r}{1.00}\\
    \multicolumn{1}{l}{$T_{ma}$} &
  \multicolumn{1}{l}{RTU mixed air temperature} &
  \multicolumn{1}{l}{$^\circ \text{C}$}&
  \multicolumn{1}{r}{39.27}&
  \multicolumn{1}{r}{7.00}&
  \multicolumn{1}{r}{1.00}\\
      \multicolumn{1}{l}{$T_{oa}$} &
  \multicolumn{1}{l}{RTU outdoor air temperature} &
  \multicolumn{1}{l}{$^\circ \text{C}$}&
  \multicolumn{1}{r}{45.16}&
  \multicolumn{1}{r}{-1.00}&
  \multicolumn{1}{r}{1.00}\\
        \multicolumn{1}{l}{$\dot{V}_{sa}$} &
  \multicolumn{1}{l}{RTU filtered supply air volume flow rate} &
  \multicolumn{1}{l}{$\text{m}^\text{3} / \text{s}$}&
  \multicolumn{1}{r}{10.26}&
  \multicolumn{1}{r}{0.00}&
  \multicolumn{1}{r}{1.00}\\
            \multicolumn{1}{l}{$T_{shw}$} &
  \multicolumn{1}{l}{HP supply hot water temperature} &
  \multicolumn{1}{l}{$^\circ \text{C}$}&
  \multicolumn{1}{r}{54.77}&
  \multicolumn{1}{r}{-1.65}&
  \multicolumn{1}{r}{5.00}\\
        \multicolumn{1}{l}{$T_{rhw}$} &
  \multicolumn{1}{l}{HP return hot water temperature} &
  \multicolumn{1}{l}{$^\circ \text{C}$}&
  \multicolumn{1}{r}{53.71}&
  \multicolumn{1}{r}{-1.08}&
  \multicolumn{1}{r}{5.00}\\
            \multicolumn{1}{l}{$\dot{V}_{shw}$} &
  \multicolumn{1}{l}{HP supply hot water volume flow rate} &
  \multicolumn{1}{l}{$\text{m}^\text{3} / \text{h}$}&
  \multicolumn{1}{r}{15.44}&
  \multicolumn{1}{r}{0.01}&
  \multicolumn{1}{r}{5.00}\\
  \toprule
    \multicolumn{1}{l}{$\dot{Q}_{cool_{tot}}$} &
  \multicolumn{1}{l}{Total cooling flow rate} &
  \multicolumn{1}{l}{kW}&
  \multicolumn{1}{r}{320.13}&
  \multicolumn{1}{r}{-48.51}&
  \multicolumn{1}{r}{30.00}\\
    \multicolumn{1}{l}{$\dot{Q}_{hw}$} &
  \multicolumn{1}{l}{Reheat water flow rate at the HP} &
  \multicolumn{1}{l}{kW}&
  \multicolumn{1}{r}{116.81}&
  \multicolumn{1}{r}{-12.26}&
  \multicolumn{1}{r}{30.00}\\
    \multicolumn{1}{l}{$T_{ra_{avg}}$} &
  \multicolumn{1}{l}{Average indoor air temperature} &
  \multicolumn{1}{l}{$^\circ \text{C}$}&
  \multicolumn{1}{r}{36.59}&
  \multicolumn{1}{r}{9.53}&
  \multicolumn{1}{r}{30.00}\\
      \multicolumn{1}{l}{$T_{oa_{avg}}$} &
  \multicolumn{1}{l}{Average outdoor air temperature} &
  \multicolumn{1}{l}{$^\circ \text{C}$}&
  \multicolumn{1}{r}{43.19}&
  \multicolumn{1}{r}{2.88}&
  \multicolumn{1}{r}{30.00}\\

   \toprule  
\end{tabular}

\end{table}

Since the cooling and heating flow rates were not available, these are derived from the HVAC operational data of Table \ref{tab:hvac}, as follows:

\begin{flalign}
&\dot{Q}_{cool_{i}} = \dot{V}_{sa_{i}} \cdot \rho_{a} \cdot c_{p_{a}} \cdot (T_{sa_{i}}-T_{ma_{i}}) ~,&
\\
&\dot{Q}_{heat} = f \cdot \dot{V}_{hw} \cdot \rho_{w} \cdot c_{p_{w}} \cdot (T_{shw}-T_{rhw}) = f \cdot \dot{Q}_{hw}~,&
\label{eq:q}
\end{flalign}

where $\dot{Q}_{cool_{i}}$ ([W]) is the cooling flow rate from the i-th RTU, $\dot{V}_{sa_{i}} $ ([$\frac{\text{m}^\text{3}}{\text{s}}$]) is the i-th RTU filtered supply air volume flow rate, $\rho_{a}$ and  $\rho_{w}$ are the densities of the air (1.204 $\frac{\text{kg}}{\text{m}^\text{3}}$) and water (1000 $\frac{\text{kg}}{\text{m}^\text{3}}$), $c_{p_{a}}$ and  $c_{p_{w}}$ are the specific heat capacities of the air (1006 $\frac{\text{J}}{\text{kg}\cdot \text{K}}$) and water (4200 $\frac{\text{J}}{\text{kg}\cdot \text{K}}$), $T_{sa_{i}}$ ([K]) is the i-th RTU supply air temperature, $T_{ma_{i}}$ ([K]) is the i-th RTU mixed air temperature, $f$ ([-]) is an unknown correction factor,  $\dot{V}_{shw}$ ([$\frac{\text{m}^\text{3}}{\text{s}}$]) is the HP supply hot water volume flow rate,  ${T}_{shw}$ ([K]) is the HP supply hot water temperature, ${T}_{rhw}$ ([K]) is the HP return hot water temperature  and $\dot{Q}_{hw}$ ([W]) is the reheat water heat flow rate.

Each UTF serves a specific thermal zone in the external perimeter. However, as confirmed by Blum et al. \cite{blum2022field}, it is difficult to quantify how much air is flowing in each zone. For simplicity, data are therefore aggregated at the building level. Following Blum et al. \cite{blum2022field}, the average indoor air temperature ($T_{ra_{avg}}$) and average outdoor air temperature ($T_{oa_{avg}}$) data are obtained by the mean operation, while the total cooling flow rate ($\dot{Q}_{cool_{tot}}$) is obtained by summing the contribution of each RTU. Here, assuming perfect mixing of the air, the indoor air temperature of the thermal zone served by the i-th RTU is considered equal to the i-th RTU return air temperature $T_{ra_{i}}$. Finally, in line with the previous studies from which the Autoencoders are adopted \cite{liguori2021indoor}, the dataset is resampled to a 30 minutes-based frequency. The final dataset can be observed in the lower part of Table \ref{tab:hvac}. Since the cooling and heating flow rates are computed with the HVAC operational data of the upper part of the Table, the occurrence of negative values is attributed to the respective negative HVAC operational data. This might indicate that sensors' calibration is needed \cite{blum2022field}. For additional information, the reader is referred to Blum et al. \cite{blum2022field}.

\subsection{Building thermal balance}



In order to implement a Physics-informed Neural Network, a PDE that guides the outputs towards physically meaningful values has to be determined. As shown by Bertagnolio and Lebrun  \cite{bertagnolio2008simulation}, several equations can be used to describe the energy and mass balances in a building. For instance, equations governing the sensible heat, $CO_2$, and water balances can be established. In the scope of this paper it is aimed to impute indoor air temperature, heating, and cooling data. This might be particularly valuable for improved retrofit analysis. As observed by Ferrari and Zanotto \cite{ferrari2015building}, these variables are primarily governed by the following ODE:

\begin{flalign}
&\dot{Q}_{storage} = M\cdot c_{p_{a}} \cdot \frac{dT_{ra_{avg}}}{dt} = \dot{Q}_{env} + \dot{Q}_{ven} + \dot{Q}_{int} + \dot{Q}_{sol} - \notag\\
&\dot{Q}_{cool_{tot}} + \dot{Q}_{heat}~,&
\label{eq:thermal_balance}
\end{flalign}

where $M$ is the air volume mass ([kg]). In the presented case study, it is difficult to obtain correct estimations for the ventilation ($\dot{Q}_{ven}$), internal ($\dot{Q}_{int}$) and solar ($\dot{Q}_{sol}$) heat flow rates. Therefore, in line with previous works in literature \cite{gokhale2022physics,di2022physically}, these contributions are ignored. The heat flow rate from the environment ($\dot{Q}_{env}$) is defined as follows:

\begin{flalign}
&\dot{Q}_{env} = A \cdot U_{w} \cdot (T_{oa_{avg}} - T_{ra_{avg}})~,&
\\
& U_w = \frac{\sum_{i}A_i \cdot U_i}{\sum_{i}A_i}~,&
\label{eq:env}
\end{flalign}

where $A$ ([$\text{m}^\text{2}$]) is the total surface area of the building and $U_w$ ([$\frac{\text{W}}{\text{m}^\text{2} \cdot \text{K}}$]) is a weighted heat transmission coefficient over the $i-th$ external surface area of the building. By rearranging all the terms, Equation \ref{eq:final} is obtained:

\begin{flalign}
&\frac{dT_{ra_{avg}}}{dt} = a_1 \cdot (T_{oa_{avg}} - T_{ra_{avg}}) - b_1 \cdot \dot{Q}_{cool_{tot}} + c_1 \cdot \dot{Q}_{hw}~,&
\label{eq:final}
\end{flalign}

where $a_1$ ([$\frac{\text{1}}{\text{s}}$]), $b_1$ ([$\frac{\text{K}}{\text{J}}$]) and $c_1$ ([$\frac{\text{K}}{\text{J}}$]) are unknown physics-based parameters that should be greater than zero to be physically meaningful. By approximating Equation \ref{eq:final} with the forward difference method \cite{van2010introduction}, a final formulation which is true for each timestep $t$ is obtained:

\begin{flalign}
&(T_{ra_{avg_{t+1}}} - T_{ra_{avg_{t}}}) - ( a \cdot (T_{oa_{avg_{t}}} - T_{ra_{avg_{t}}}) - b \cdot \dot{Q}_{cool_{tot_{t}}} + \notag\\
&c \cdot \dot{Q}_{hw_{t}} ) = 0~, \forall t~,&
\label{eq:final_fw}
\end{flalign}

where the fixed interval of time $\Delta t$ is included in the new unknown physics-based parameters, i.e. $a$ ([-]), $b$ ([$\frac{\text{K}}{\text{W}}$]) and $c$ ([$\frac{\text{K}}{\text{W}}$]).
\subsection{From DAE to PI-DAE}

\begin{figure}[ht]
\centering
\includegraphics[width=1.\textwidth]{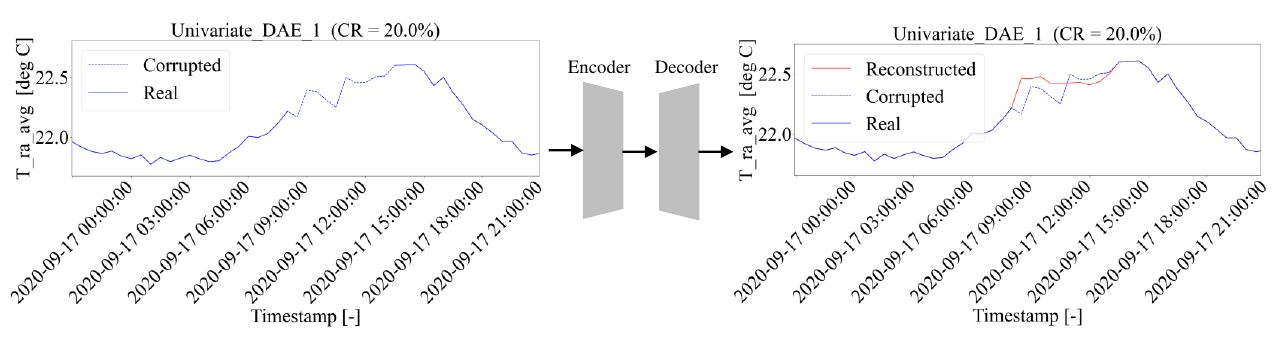}
\caption{Working principle of a generic univariate DAE for missing indoor air temperature data imputation with 20\% corruption rate (CR).}
\label{fig:autoencoder}
\end{figure}

A Denoising Autoencoder is a representation learning approach that is trained to reconstruct a corrupted input, i.e., a vector with noisy elements \cite{goodfellow2016deep}. In the case of missing data imputation problems, noisy elements are missing values. A generic representation of a univariate DAE for the reconstruction of daily profiles is shown in Figure \ref{fig:autoencoder}. A single day of observation of indoor air temperature data is corrupted by replacing some of its elements with zeros. The sample is compressed to a latent space vector through an encoder and decompressed through a decoder. In the decompressed sample, the corrupted elements have been imputed according to the learned probability distribution \cite{liguori2023augmenting}. 

The selected univariate DAE is a Convolutional Denoising Autoencoder implemented as part of Liguori et al. \cite{liguori2021indoor}. It was already proven that the aforementioned Denoising Autoencoder could successfully generalize to alternative buildings \cite{liguori2021gap,liguori2023augmenting}. For that reason, in this paper, the network architecture is preserved while just some hyperparameters are optimized. Further details about the model optimization are presented in Section \ref{subsec: development}. Eventually, three different univariate DAEs for missing indoor air temperature (Univariate\_DAE\_1), heating (Univariate\_\_DAE\_2), and cooling (Univariate\_\_DAE\_3) data imputation are optimized. Additionally, two different multivariate DAEs, where the missing intervals for each input time series are assumed to occur at the same time, are implemented and optimized based on the same procedure. Namely, Multivariate\_DAE\_1 takes as input corrupted indoor air temperature, heating, and cooling data. Multivariate\_DAE\_2 takes as input corrupted indoor air temperature, heating, cooling, and full outdoor air temperature data. Here, since weather data can usually be obtained from nearby weather stations, the outdoor air temperature is not corrupted.

\begin{figure}[ht]
\centering
\includegraphics[width=1.\textwidth]{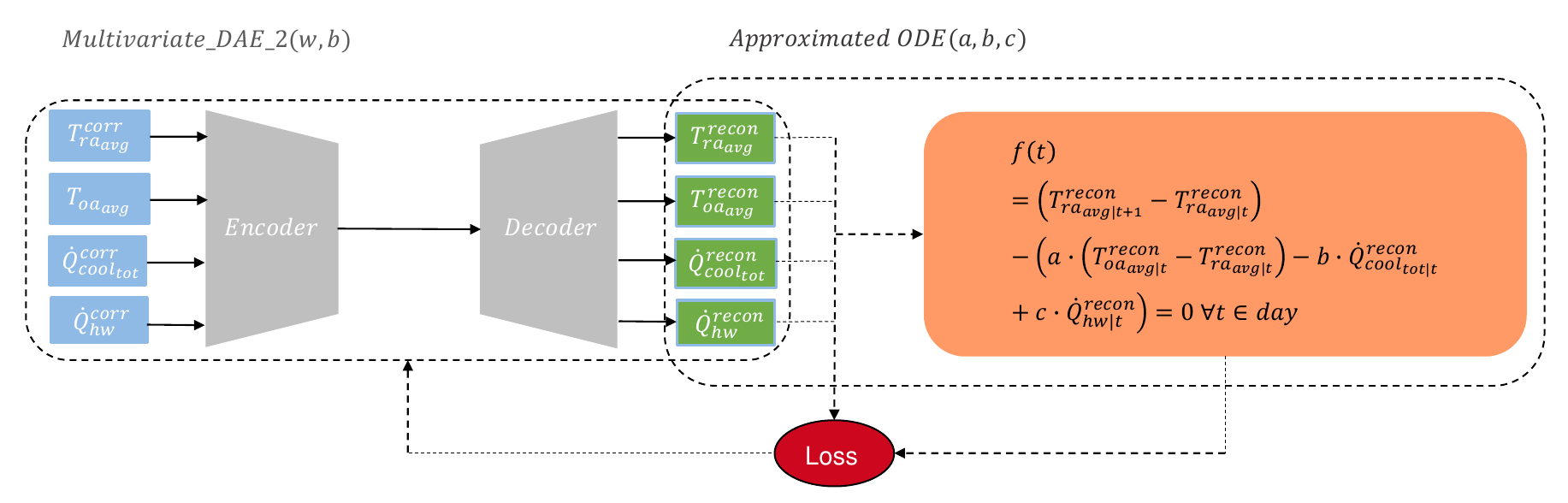}
\caption{Representation of the proposed PI-DAE with approximated building energy balance ODE. The analyzed variables are the total cooling flow rate ($\dot{Q}_{cool_{tot}}$), the reheat water flow rate ($\dot{Q}_{hw}$), the average indoor air temperature  ($T_{ra_{avg}}$) and the average outdoor air temperature ($T_{oa_{avg}}$). Figure partially reproduced based on Jagtap et al. \cite{jagtap2020adaptive}.}
\label{fig:pidae}
\end{figure}

The proposed PI-DAE can be observed in Figure \ref{fig:pidae}. It consists of two parts, namely a multivariate DAE with outdoor air temperature (Multivariate\_DAE\_2) and an approximated ordinary differential equation based on Equation \ref{eq:final_fw} that ties the reconstructed outputs together (Approximated ODE). Here, the Approximated ODE depends on the unknown physics-based parameters that are optimized together with the weights ($w$) and biases ($b$) of the DAE component. In particular, in order to quantify the effects of the physical component on the Autoencoder, the hyperparameters of PI-DAE are the same as those of Multivariate\_DAE\_2.  

The total loss function of the model ($Loss$) is therefore given by a regression loss ($Loss_{Mutlivariate\_DAE\_2}$) and a physics-based loss ($Loss_{Approximated ODE}$), as follows \cite{jagtap2020adaptive}:

\begin{flalign}
&Loss = Loss_{Mutlivariate\_DAE\_2} + Loss_{Approximated ODE}~,&
\label{eq:loss}
\end{flalign}

where both losses are given the same weight to assess the impact of the physical component.





\subsection{Model optimization}
\label{subsec: development}

The adopted Convolutional Denoising Autoencoder is a symmetric Autoencoder with two convolutional layers both in the encoder (with down-sampling) and decoder (with up-sampling). A detailed overview of the aforementioned model architecture is presented in the Appendix of Liguori et al. \cite{liguori2023augmenting}. As explained earlier, the network architecture is the same for every DAE configuration. However, the hyperparameter tuning is performed separately for every model and corruption rate (CR). A summary of the tuned hyperparameters is presented in Table \ref{tab:hyperparameter} (top part). Tuning is conducted using Optuna \cite{akiba2019optuna}, with a number of trials equal to 200.  Further information regarding the tuned hyperparameters is provided in the open-source repository\footnote{\url{https://github.com/Antonio955/PI_DAE.git}\label{footnote 1}}. 

\begin{table}[ht]
\centering
\caption{Overview of the tuned hyperparameters and average number of trainable parameters.}
\label{tab:hyperparameter}

\begin{tabular}{ll}\\
      \toprule
  \multicolumn{1}{l}{Hyperparameter} & \multicolumn{1}{r}{Interval}\\
\toprule
    \multicolumn{1}{l}{Filters external layers} &
  \multicolumn{1}{r}{5 - 200}\\
    \multicolumn{1}{l}{Filters internal layers} &
  \multicolumn{1}{r}{5 - 200}\\
      \multicolumn{1}{l}{Kernel size} &
  \multicolumn{1}{r}{1 - 10}\\
        \multicolumn{1}{l}{Learning rate} &
  \multicolumn{1}{r}{0.0001 - 0.1}\\
          \multicolumn{1}{l}{Batch size} &
  \multicolumn{1}{r}{32 - 256}\\
   \toprule  
          \multicolumn{1}{l}{Model} &
  \multicolumn{1}{r}{Average number of trainable parameters} \\
 \toprule  
          \multicolumn{1}{l}{Univariate\_DAE\_1} &
  \multicolumn{1}{r}{73462.25}\\
          \multicolumn{1}{l}{Univariate\_DAE\_2} &
  \multicolumn{1}{r}{33980.50}\\
          \multicolumn{1}{l}{Univariate\_DAE\_3} &
  \multicolumn{1}{r}{47644.00}\\
           \multicolumn{1}{l}{Multivariate\_DAE\_1} &
  \multicolumn{1}{r}{69059.50}\\
           \multicolumn{1}{l}{Multvariate\_DAE\_2} &
  \multicolumn{1}{r}{8919.00}\\
            \multicolumn{1}{l}{PI-DAE} &
  \multicolumn{1}{r}{8922.00}\\
    \toprule  
\end{tabular}
\end{table}

Table \ref{tab:hyperparameter} also shows the average number of trainable parameters for each optimized model on the bottom part. After the tuning process, each Autoencoder has a different level of complexity. Specifically, the level of complexity of PI-DAE and Multivariate\_DAE\_2, in terms of the number of trainable parameters, is one order of magnitude less than all the remaining DAE configurations. Furthermore, PI-DAE has just three additional trainable parameters compared to Multivariate\_DAE\_2. In this regard, although the hyperparameters of the two models are the same, it is noted that the physics-based parameters of PI-DAE need to be optimized during training. 

Finally, in order to prevent overfitting on small datasets, each model implements early stopping criteria \cite{goodfellow2016deep}, and training data are augmented with masking noise injection \cite{liguori2023augmenting}.  

\subsection{Building monitoring periods}
\label{subsec:monitoring}

Embedding physical knowledge in a deep learning model foresees the input and output features being tied by some specific mathematical expression. In the scope of this paper, it is assumed that the reconstructed variables follow the behavior represented by Equation \ref{eq:final_fw}. However, following the same procedure, the proposed multivariate DAE might be coupled with any expression enforcing physical laws on the outputs. 

In the presented study, Equation \ref{eq:final_fw} depends on strong hypotheses, e.g., the absence of solar radiation, which might lead to biased and inaccurate results. A more precise way to represent the building's thermal dynamics would be to account for external disturbances, such as weather conditions and occupant behavior, with an additional noise term \cite{andersen2000modelling}. On the one hand, solving such a stochastic ODE requires a much more complex model due to the so-called ``curse of dimensionality" \cite{karniadakis2021physics}. On the other hand, the absence of the noise term might force the model to exploit the physics-based coefficients to account for the unmodeled disturbances \cite{brastein2018parameter}. In this paper, the focus is on the effectiveness of PI-DAE in detecting building monitoring periods with unmeasured phenomena through the learned physics-based coefficients. 

In order to evaluate such a capability, the different building monitoring periods should first be extracted from the full dataset. It is argued that if the effects of the external disturbances on the indoor air temperature are really small, then the considered variables should be highly correlated with each other. For instance, a low correlation between the cooling flow rate and the indoor air temperature could indicate that other factors, such as occupant behavior, should be taken into account. However, this hypothesis must be taken carefully, as correlation is a measure of association and not of casual relationship \cite{hung2017interpretation}. In this sense, coupling this analysis with the learned physics-based coefficients might provide building practitioners with important physical insights about the analyzed dataset.

In this work, the Pearson Correlation Coefficient (PCC) is used to quantify the correlation attributes since it is widely adopted in the literature \cite{qureshi2017wind}. Considering that higher PCCs are observed on days with higher heating and cooling flow rate variability, data are filtered based on the interquartile range (IQR) of these two variables. Namely, only days over a certain threshold are selected. However, the selection procedure of the highly correlated days might depend on the particular building, and a more comprehensive investigation is out of the scope of this study. The use of the IQR as a measure of data variability is based on its wide application in the literature \cite{ahmad2012efficient}.

Table \ref{tab:correlation} shows the obtained Pearson Correlation Coefficients among the analyzed variables by selecting different IQR thresholds for the cooling and heating flow rates. Here, several combinations between zero and 50 kW are analyzed by using a step size of 10 kW. No significant difference was noted by using a smaller step size. Specifically, only the combinations for which the resulting dataset has a number of days bigger than zero are represented. As earlier discussed, the correlation coefficients are highly dynamic due to the varying unsensed disturbances over time. This explains the marked variability of the PCCs under different monitoring periods in Table \ref{tab:correlation}. It was also observed that higher thresholds produce extremely small datasets. The definitions of the introduced coefficients are provided below. $PCC_1$ corresponds to the correlation coefficient between the indoor air temperature and cooling flow rate; $PCC_2$ corresponds to the correlation coefficient between the indoor air temperature and the heating flow rate; $PCC_3$ corresponds to the correlation coefficient between the heating flow rate and cooling flow rate; $PCC_4$ corresponds to the correlation coefficient between the outdoor air temperature and cooling flow rate; $PCC_5$ corresponds to the correlation coefficient between the outdoor air temperature and heating flow rate; $PCC_6$ corresponds to the correlation coefficient between the outdoor air temperature and indoor air temperature. It is observed that most of the highest correlation coefficients fall in the case where the IQR thresholds of the cooling and heating flow rates are 50 and 20 kW, respectively. Based on this consideration, the models' evaluation is performed on two different building operation periods. Namely, Case 1 is the full dataset, including 363 days of observations. Case 2 is the reduced dataset comprising 19 days of observation with the highest correlation among the observed variables.


\begin{table}[ht]
\centering
\caption{Pearson Correlation Coefficients (PCC) with different interquartile range (IQR) thresholds for the cooling ($\dot{Q}_{cool_{tot}}$) and heating ($\dot{Q}_{hw}$) flow rates. $Days$ is the number of days of the filtered dataset; $PCC_1$ corresponds to the correlation coefficient between the indoor air temperature and cooling flow rate; $PCC_2$ corresponds to the correlation coefficient between the indoor air temperature and the heating flow rate; $PCC_3$ corresponds to the correlation coefficient between the heating flow rate and cooling flow rate; $PCC_4$ corresponds to the correlation coefficient between the outdoor air temperature and cooling flow rate; $PCC_5$ corresponds to the correlation coefficient between the outdoor air temperature and heating flow rate; $PCC_6$ corresponds to the correlation coefficient between the outdoor air temperature and indoor air temperature.
}
\label{tab:correlation}

\begin{tabular}{lllllllll}\\
      \toprule
\multicolumn{1}{l}{$IQR(\dot{Q}_{cool_{tot}})$ [kW]} &
  \multicolumn{1}{l}{$IQR(\dot{Q}_{hw})$ [kW]} &
  \multicolumn{1}{l}{$Days$} &
  \multicolumn{1}{l}{$PCC_1$} &
  \multicolumn{1}{l}{$PCC_2$} &
  \multicolumn{1}{l}{$PCC_3$} &
  \multicolumn{1}{l}{$PCC_4$} &
  \multicolumn{1}{l}{$PCC_5$} &
  \multicolumn{1}{l}{$PCC_6$} \\
     \toprule

0  & 0  & 363 & 0.2959          & -0.0830          & -0.3580          & 0.7290          & -0.4878          & 0.5231          \\
0  & 10 & 363 & 0.2959          & -0.0830          & -0.3580          & 0.7290          & -0.4878          & 0.5231          \\
0  & 20 & 363 & 0.2959          & -0.0830          & -0.3580          & 0.7290          & -0.4878          & 0.5231          \\
0  & 30 & 363 & 0.2959          & -0.0830          & -0.3580          & 0.7290          & -0.4878          & 0.5231          \\
0  & 40 & 363 & 0.2959          & -0.0830          & -0.3580          & 0.7290          & -0.4878          & 0.5231          \\
0  & 50 & 363 & 0.2959          & -0.0830          & -0.3580          & 0.7290          & -0.4878          & 0.5231          \\
10 & 0  & 363 & 0.2959          & -0.0830          & -0.3580          & 0.7290          & -0.4878          & 0.5231          \\
10 & 10 & 99  & 0.2816          & -0.5050          & -0.4619          & 0.7123          & -0.5987          & 0.5575          \\
10 & 20 & 54  & 0.3332          & -0.4410          & -0.5186          & 0.6827          & -0.6703          & 0.6184          \\
10 & 30 & 24  & 0.3151          & -0.3543          & -0.4883          & 0.6653          & -0.7015          & 0.5917          \\
20 & 0  & 363 & 0.2959          & -0.0830          & -0.3580          & 0.7290          & -0.4878          & 0.5231          \\
20 & 10 & 64  & 0.3325          & -0.3494          & -0.5214          & 0.7749          & -0.5584          & 0.5190          \\
20 & 20 & 32  & 0.4046          & -0.3333          & -0.5587          & 0.7472          & -0.6634          & 0.6122          \\
20 & 30 & 14  & 0.3331          & -0.2217          & -0.5250          & 0.7373          & -0.7012          & 0.5366          \\
30 & 0  & 363 & 0.2959          & -0.0830          & -0.3580          & 0.7290          & -0.4878          & 0.5231          \\
30 & 10 & 54  & 0.3510          & -0.3337          & -0.5255          & 0.7991          & -0.5612          & 0.5041          \\
30 & 20 & 28  & 0.4226          & -0.3215          & -0.5688          & 0.7727          & -0.6806          & 0.5921          \\
30 & 30 & 11  & 0.3744          & -0.2043          & -0.5460          & 0.7802          & \textbf{-0.7454} & 0.5057          \\
40 & 0  & 363 & 0.2959          & -0.0830          & -0.3580          & 0.7290          & -0.4878          & 0.5231          \\
40 & 10 & 48  & 0.3728          & -0.4174          & -0.5384          & 0.8123          & -0.5496          & 0.5358          \\
40 & 20 & 22  & 0.5605          & -0.5672          & \textbf{-0.6136} & 0.7999          & -0.7033          & 0.7688          \\
50 & 0  & 363 & 0.2959          & -0.0830          & -0.3580          & 0.7290          & -0.4878          & 0.5231          \\
50 & 10 & 44  & 0.3815          & -0.4286          & -0.5338          & 0.8226          & -0.5544          & 0.5300          \\
50 & 20 & 19  & \textbf{0.6154} & \textbf{-0.6132} & -0.6094          & \textbf{0.8257} & -0.7317          & \textbf{0.7822}\\
      \toprule
      
\end{tabular}
\end{table}

\subsection{Experimental design}
\label{sec:design}

In order to evaluate the performance of PI-DAE and to determine the impact of the physics-informed loss function, an ablation study is performed among different DAE configurations. Considering the stochastic initialization of the models' weights, Autoencoders are trained 10 times on the training set and validated on the validation set. The models with the lowest validation errors are further exported for evaluation. Additionally, two commonly used imputation benchmarks, namely Linear Interpolation (LIN) and k-nearest Neighbors Impute (KNN) \cite{liguori2023augmenting}, are used to further support the results. In particular, these models are also applied to impute sub-daily missing values in the same paper describing the analyzed dataset \cite{luo2022three}, hence further motivating the use of this study.  

To analyze the influence of the number of training data on the models' performance, the training rate is varied from 10 \% (36 days for Case 1 and one day for Case 2) to 50\% (181 days for Case 1 and 9 days for Case 2). This implies that, for each training rate, the models undergo a complete re-training on reduced sizes of training sets. In this case, in line with the previous studies \cite{liguori2021indoor,liguori2023augmenting}, the validation rate is set at 10\% while the rest is left for evaluation. To obtain unbiased training, validation, and evaluation sets, each of these is randomly selected 10 times from the dataset, and results are averaged. 

Finally, as continuous missing scenarios are usually more complex than random ones \cite{liguori2023augmenting}, corrupted elements are selected based on the former. In particular, the first missing entry of each corrupted day is randomly given by a discrete uniform distribution, while the rest is obtained with forward expansion. The reader is referred to Liguori et al. \cite{liguori2023augmenting} for additional information about the used continuous missing masking noise pseudo-algorithm. The daily corruption rate varies between a few hours (20\%) and around 20 hours (80\%). For reference, the trained models, source codes, and performed experiments are published as an open-source repository$^{\ref{footnote 1}}$.

\section{Experiments}
\label{sec:results}
\begin{figure}[ht]
\centering
\includegraphics[width=1\textwidth]{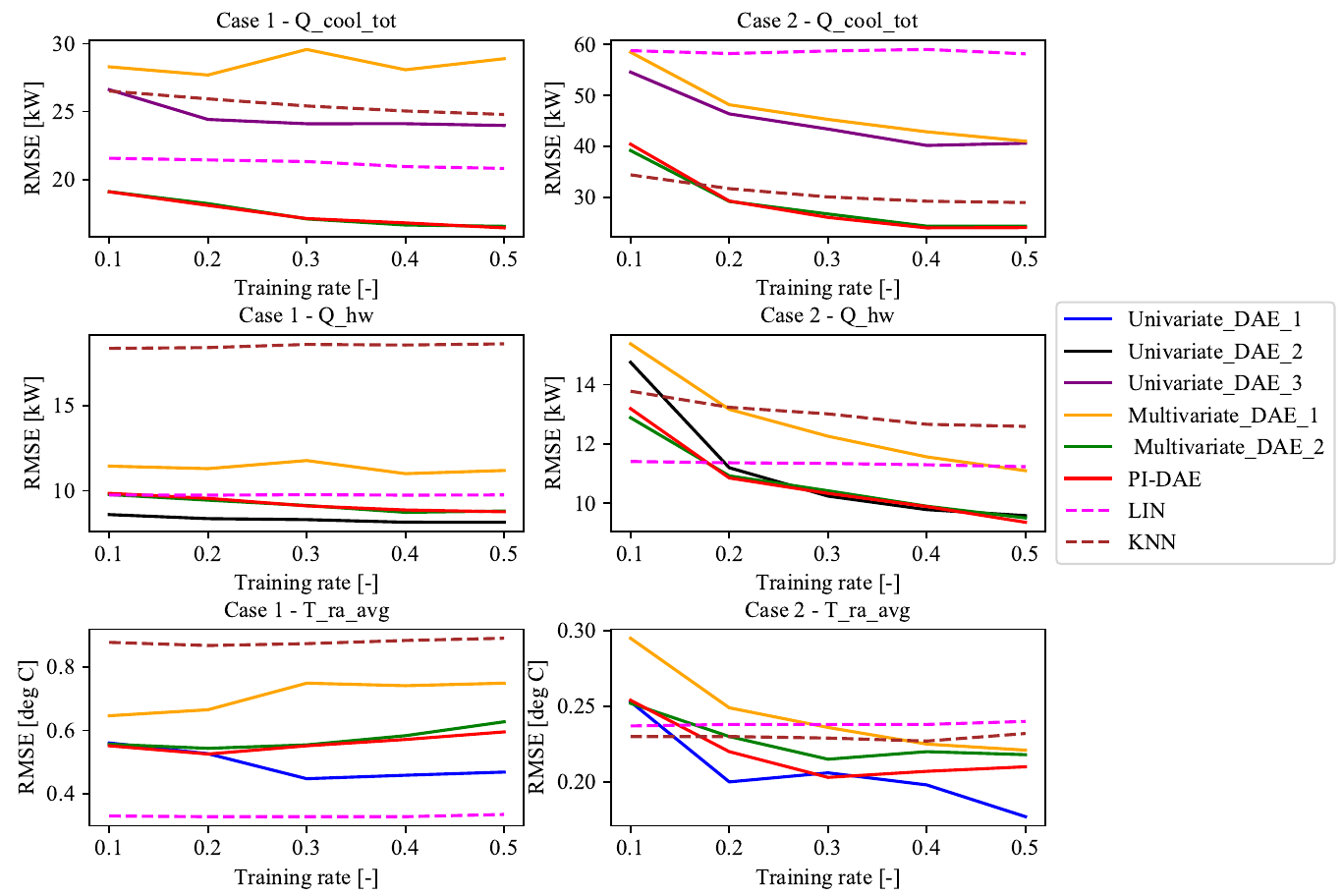}
\caption{Learning curves of the models at varied training rates and cases. For Case 1, the depicted training rates (from 0.1 to 0.5) correspond to training set sizes of 36, 72, 108, 145, and 181 days, respectively. Alternatively, for Case 2, the depicted training rates (from 0.1 to 0.5) correspond to training set sizes of 1, 3, 5, 7, and 9 days, respectively. The analyzed variables are the total cooling flow rate ($\dot{Q}_{cool_{tot}}$), the reheat water flow rate ($\dot{Q}_{hw}$) and the average indoor air temperature ($T_{ra_{avg}}$).}
\label{fig:results_1}
\end{figure}

The learning curves for the analyzed models can be observed in Figure \ref{fig:results_1}, where the root mean squared error (RMSE) has been averaged over different corruption rates and days of observation. Additionally, the same reconstruction errors are presented in detail in Table \ref{tab:results_1}. For better interpretability of the results, the Linear Interpolation model is selected as the main baseline. Additionally, the percentage decrease of the RMSE with respect to the latter is shown in parenthesis. 

\begin{sidewaystable}
\centering
\caption{Reconstruction errors of the analyzed models at different training rates (TR). Here, the training rate is defined as the number of training data over the total number of data. For Case 1, the depicted training rates (from 10 to 50\%) correspond to training set sizes of 36, 72, 108, 145, and 181 days, respectively. Alternatively, for Case 2, the depicted training rates (from 10 to 50\%) correspond to training set sizes of 1, 3, 5, 7, and 9 days, respectively. The percentage decrease with respect to the Linear Interpolation model (LIN) is shown in parenthesis. The analyzed variables are the total cooling flow rate ($\dot{Q}_{cool_{tot}}$), the reheat water flow rate ($\dot{Q}_{hw}$) and the average indoor air temperature ($T_{ra_{avg}}$).}
\label{tab:results_1}
\begin{tabular}{llllllll}\\
      \toprule
  \multicolumn{1}{l}{TR} & \multicolumn{1}{l}{Model} &
  \multicolumn{3}{l}{RMSE (Case 1)} & \multicolumn{3}{l}{RMSE (Case 2)}\\  
  \cmidrule{3-8}
  \multicolumn{1}{l}{} &
  \multicolumn{1}{l}{} &
  \multicolumn{1}{l}{$\dot{Q}_{cool_{tot}}$ [kW]}  & \multicolumn{1}{l}{$\dot{Q}_{hw}$ [kW]}  & \multicolumn{1}{l}{$T_{ra_{avg}}[^{\circ}\text{C}]$} & 
  \multicolumn{1}{l}{$\dot{Q}_{cool_{tot}}$ [kW]}  & \multicolumn{1}{l}{$\dot{Q}_{hw}$ [kW]}  & \multicolumn{1}{l}{$T_{ra_{avg}}[^{\circ}\text{C}]$}\\
\toprule
    \multicolumn{1}{l}{10\%} &
  \multicolumn{1}{l}{LIN} &
  \multicolumn{1}{l}{21.563}&
  \multicolumn{1}{l}{9.753}&
  \multicolumn{1}{l}{\textbf{0.329}}&
  \multicolumn{1}{l}{58.770}&
  \multicolumn{1}{l}{\textbf{11.410}}&
  \multicolumn{1}{l}{0.237}\\
      \multicolumn{1}{l}{} &
  \multicolumn{1}{l}{KNN} &
  \multicolumn{1}{l}{26.513 (+22.9\%)}&
  \multicolumn{1}{l}{18.355 (+88.2\%)}&
  \multicolumn{1}{l}{0.878 (+167.1\%)}&
  \multicolumn{1}{l}{\textbf{34.395 (-41.5\%)}}&
  \multicolumn{1}{l}{13.777 (+20.7\%)}&
  \multicolumn{1}{l}{\textbf{0.230 (-3.0\%)}}\\
      \multicolumn{1}{l}{} &
  \multicolumn{1}{l}{Univariate\_DAE} &
  \multicolumn{1}{l}{26.611 (+23.4\%)}&
  \multicolumn{1}{l}{\textbf{8.601 (-11.8\%)}}&
  \multicolumn{1}{l}{0.560 (+70.4\%)}&
  \multicolumn{1}{l}{54.560 (-7.2\%)}&
  \multicolumn{1}{l}{14.750 (+29.3\%)}&
  \multicolumn{1}{l}{0.253 (+6.8\%)}\\
      \multicolumn{1}{l}{} &
  \multicolumn{1}{l}{Multivariate\_DAE\_1} &
  \multicolumn{1}{l}{28.294 (+31.2\%)}&
  \multicolumn{1}{l}{11.445 (+17.3\%)}&
  \multicolumn{1}{l}{0.646 (+96.5\%)}&
  \multicolumn{1}{l}{58.469 (-0.5\%)}&
  \multicolumn{1}{l}{15.372 (+34.7\%)}&
  \multicolumn{1}{l}{0.295 (+24.6\%)}\\
      \multicolumn{1}{l}{} &
  \multicolumn{1}{l}{Multivariate\_DAE\_2} &
  \multicolumn{1}{l}{19.099 (-11.4\%)}&
  \multicolumn{1}{l}{9.777 (+0.2\%)}&
  \multicolumn{1}{l}{0.556 (+69.2\%)}&
  \multicolumn{1}{l}{39.166 (-33.4\%)}&
  \multicolumn{1}{l}{12.884 (+12.9\%)}&
  \multicolumn{1}{l}{0.252 (+6.1\%)}\\
      \multicolumn{1}{l}{} &
  \multicolumn{1}{l}{PI-DAE} &
  \multicolumn{1}{l}{\textbf{19.093 (-11.5\%)}}&
  \multicolumn{1}{l}{9.849 (+1.0\%)}&
  \multicolumn{1}{l}{0.551 (+67.8\%)}&
  \multicolumn{1}{l}{40.415 (-31.2\%)}&
  \multicolumn{1}{l}{13.191 (+15.6\%)}&
  \multicolumn{1}{l}{0.254 (+7.3\%)}\\
   \toprule  
    \multicolumn{1}{l}{20\%} &
  \multicolumn{1}{l}{LIN} &
  \multicolumn{1}{l}{21.437}&
  \multicolumn{1}{l}{9.741}&
  \multicolumn{1}{l}{\textbf{0.327}}&
  \multicolumn{1}{l}{58.206}&
  \multicolumn{1}{l}{11.366}&
  \multicolumn{1}{l}{0.238}\\
      \multicolumn{1}{l}{} &
  \multicolumn{1}{l}{KNN} &
  \multicolumn{1}{l}{25.935 (21.0\%)}&
  \multicolumn{1}{l}{18.401 (88.9\%)}&
  \multicolumn{1}{l}{0.868 (165.0\%)}&
  \multicolumn{1}{l}{31.691 (-45.6\%)}&
  \multicolumn{1}{l}{13.234 (16.4\%)}&
  \multicolumn{1}{l}{0.230 (-3.3\%)}\\
      \multicolumn{1}{l}{} &
  \multicolumn{1}{l}{Univariate\_DAE} &
  \multicolumn{1}{l}{24.419 (13.9\%)}&
  \multicolumn{1}{l}{\textbf{8.364 (-14.1\%)}}&
  \multicolumn{1}{l}{0.526 (60.8\%)}&
  \multicolumn{1}{l}{46.361 (-20.4\%)}&
  \multicolumn{1}{l}{11.199 (-1.5\%)}&
  \multicolumn{1}{l}{\textbf{0.200 (-15.9\%)}}\\
      \multicolumn{1}{l}{} &
  \multicolumn{1}{l}{Multivariate\_DAE\_1} &
  \multicolumn{1}{l}{27.691 (29.2\%)}&
  \multicolumn{1}{l}{11.298 (16.0\%)}&
  \multicolumn{1}{l}{0.665 (103.2\%)}&
  \multicolumn{1}{l}{48.156 (-17.3\%)}&
  \multicolumn{1}{l}{13.166 (15.8\%)}&
  \multicolumn{1}{l}{0.249 (4.6\%)}\\
      \multicolumn{1}{l}{} &
  \multicolumn{1}{l}{Multivariate\_DAE\_2} &
  \multicolumn{1}{l}{18.226 (-15.0\%)}&
  \multicolumn{1}{l}{9.449 (-3.0\%)}&
  \multicolumn{1}{l}{0.543 (65.9\%)}&
  \multicolumn{1}{l}{\textbf{29.185 (-49.9\%)}}&
  \multicolumn{1}{l}{10.925 (-3.9\%)}&
  \multicolumn{1}{l}{0.230 (-3.5\%)}\\
      \multicolumn{1}{l}{} &
  \multicolumn{1}{l}{PI-DAE} &
  \multicolumn{1}{l}{\textbf{18.091 (-15.6\%)}}&
  \multicolumn{1}{l}{9.556 (-1.9\%)}&
  \multicolumn{1}{l}{0.525 (60.4\%)}&
  \multicolumn{1}{l}{29.256 (-49.7\%)}&
  \multicolumn{1}{l}{\textbf{10.854 (-4.5\%)}}&
  \multicolumn{1}{l}{0.220 (-7.5\%)}\\
   \toprule  
    \multicolumn{1}{l}{30\%} &
  \multicolumn{1}{l}{LIN} &
  \multicolumn{1}{l}{21.318}&
  \multicolumn{1}{l}{9.768}&
  \multicolumn{1}{l}{\textbf{0.327}}&
  \multicolumn{1}{l}{58.708}&
  \multicolumn{1}{l}{11.344}&
  \multicolumn{1}{l}{0.238}\\
      \multicolumn{1}{l}{} &
  \multicolumn{1}{l}{KNN} &
  \multicolumn{1}{l}{25.412 (+19.2\%)}&
  \multicolumn{1}{l}{18.586 (+90.3\%)}&
  \multicolumn{1}{l}{0.874 (+166.9\%)}&
  \multicolumn{1}{l}{30.064 (-48.8\%)}&
  \multicolumn{1}{l}{13.013 (+14.7\%)}&
  \multicolumn{1}{l}{0.229 (-3.9\%)}\\
      \multicolumn{1}{l}{} &
  \multicolumn{1}{l}{Univariate\_DAE} &
  \multicolumn{1}{l}{24.103 (+13.1\%)}&
  \multicolumn{1}{l}{\textbf{8.307 (-15.0\%)}}&
  \multicolumn{1}{l}{0.447 (+36.6\%)}&
  \multicolumn{1}{l}{43.381 (-26.1\%)}&
  \multicolumn{1}{l}{\textbf{10.244 (-9.7\%)}}&
  \multicolumn{1}{l}{0.206 (-13.3\%)}\\ 
      \multicolumn{1}{l}{} &
  \multicolumn{1}{l}{Multivariate\_DAE\_1} &
  \multicolumn{1}{l}{29.573 (+38.7\%)}&
  \multicolumn{1}{l}{11.773 (+20.5\%)}&
  \multicolumn{1}{l}{0.749 (+128.6\%)}&
  \multicolumn{1}{l}{45.279 (-22.9\%)}&
  \multicolumn{1}{l}{12.260 (+8.1\%)}&
  \multicolumn{1}{l}{0.236 (-0.8\%)}\\
      \multicolumn{1}{l}{} &
  \multicolumn{1}{l}{Multivariate\_DAE\_2} &
  \multicolumn{1}{l}{\textbf{17.096 (-19.8\%)}}&
  \multicolumn{1}{l}{9.136 (-6.5\%)}&
  \multicolumn{1}{l}{0.554 (+69.1\%)}&
  \multicolumn{1}{l}{26.717 (-54.5\%)}&
  \multicolumn{1}{l}{10.423 (-8.1\%)}&
  \multicolumn{1}{l}{0.215 (-9.7\%)}\\
      \multicolumn{1}{l}{} &
  \multicolumn{1}{l}{PI-DAE} &
  \multicolumn{1}{l}{17.122 (-19.7\%)}&
  \multicolumn{1}{l}{9.112 (-6.7\%)}&
  \multicolumn{1}{l}{0.551 (+68.2\%)}&
  \multicolumn{1}{l}{\textbf{26.044 (-55.6\%)}}&
  \multicolumn{1}{l}{10.335 (-8.9\%)}&
  \multicolumn{1}{l}{\textbf{0.203 (-14.9\%)}}\\
   \toprule  
    \multicolumn{1}{l}{40\%} &
  \multicolumn{1}{l}{LIN} &
  \multicolumn{1}{l}{20.947}&
  \multicolumn{1}{l}{9.743}&
  \multicolumn{1}{l}{\textbf{0.327}}&
  \multicolumn{1}{l}{59.009}&
  \multicolumn{1}{l}{11.299}&
  \multicolumn{1}{l}{0.238}\\
      \multicolumn{1}{l}{} &
  \multicolumn{1}{l}{KNN} &
  \multicolumn{1}{l}{25.042 (19.5\%)}&
  \multicolumn{1}{l}{18.552 (90.4\%)}&
  \multicolumn{1}{l}{0.884 (170.2\%)}&
  \multicolumn{1}{l}{29.217 (-50.5\%)}&
  \multicolumn{1}{l}{12.662 (12.1\%)}&
  \multicolumn{1}{l}{0.227 (-4.6\%)}\\
      \multicolumn{1}{l}{} &
  \multicolumn{1}{l}{Univariate\_DAE} &
  \multicolumn{1}{l}{24.106 (15.1\%)}&
  \multicolumn{1}{l}{\textbf{8.154 (-16.3\%)}}&
  \multicolumn{1}{l}{0.458 (40.2\%)}&
  \multicolumn{1}{l}{40.165 (-31.9\%)}&
  \multicolumn{1}{l}{\textbf{9.788 (-13.4\%)}}&
  \multicolumn{1}{l}{\textbf{0.198 (-17.0\%)}}\\
      \multicolumn{1}{l}{} &
  \multicolumn{1}{l}{Multivariate\_DAE\_1} &
  \multicolumn{1}{l}{28.076 (34.0\%)}&
  \multicolumn{1}{l}{11.004 (12.9\%)}&
  \multicolumn{1}{l}{0.741 (126.5\%)}&
  \multicolumn{1}{l}{42.845 (-27.4\%)}&
  \multicolumn{1}{l}{11.565 (2.4\%)}&
  \multicolumn{1}{l}{0.225 (-5.3\%)}\\
      \multicolumn{1}{l}{} &
  \multicolumn{1}{l}{Multivariate\_DAE\_2} &
  \multicolumn{1}{l}{\textbf{16.647 (-20.5\%)}}&
  \multicolumn{1}{l}{8.735 (-10.3\%)}&
  \multicolumn{1}{l}{0.583 (78.2\%)}&
  \multicolumn{1}{l}{24.328 (-58.8\%)}&
  \multicolumn{1}{l}{9.903 (-12.4\%)}&
  \multicolumn{1}{l}{0.220 (-7.8\%)}\\
      \multicolumn{1}{l}{} &
  \multicolumn{1}{l}{PI-DAE} &
  \multicolumn{1}{l}{16.804 (-19.8\%)}&
  \multicolumn{1}{l}{8.873 (-8.9\%)}&
  \multicolumn{1}{l}{0.571 (74.7\%)}&
  \multicolumn{1}{l}{\textbf{23.988 (-59.3\%)}}&
  \multicolumn{1}{l}{9.892 (-12.5\%)}&
  \multicolumn{1}{l}{0.207 (-13.3\%)}\\
   \toprule  
    \multicolumn{1}{l}{50\%} &
  \multicolumn{1}{l}{LIN} &
  \multicolumn{1}{l}{20.816}&
  \multicolumn{1}{l}{9.755}&
  \multicolumn{1}{l}{\textbf{0.334}}&
  \multicolumn{1}{l}{58.142}&
  \multicolumn{1}{l}{11.234}&
  \multicolumn{1}{l}{0.240}\\
      \multicolumn{1}{l}{} &
  \multicolumn{1}{l}{KNN} &
  \multicolumn{1}{l}{24.787 (+19.1\%)}&
  \multicolumn{1}{l}{18.621 (+90.9\%)}&
  \multicolumn{1}{l}{0.891 (+167.0\%)}&
  \multicolumn{1}{l}{28.961 (-50.2\%)}&
  \multicolumn{1}{l}{12.591 (+12.1\%)}&
  \multicolumn{1}{l}{0.232 (-3.5\%)}\\
      \multicolumn{1}{l}{} &
  \multicolumn{1}{l}{Univariate\_DAE} &
  \multicolumn{1}{l}{23.975 (+15.2\%)}&
  \multicolumn{1}{l}{\textbf{8.151 (-16.4\%)}}&
  \multicolumn{1}{l}{0.468 (+40.3\%)}&
  \multicolumn{1}{l}{40.600 (-30.2\%)}&
  \multicolumn{1}{l}{9.586 (-14.7\%)}&
  \multicolumn{1}{l}{\textbf{0.177 (-26.1\%)}}\\
      \multicolumn{1}{l}{} &
  \multicolumn{1}{l}{Multivariate\_DAE\_1} &
  \multicolumn{1}{l}{28.892 (+38.8\%)}&
  \multicolumn{1}{l}{11.191 (+14.7\%)}&
  \multicolumn{1}{l}{0.749 (+124.3\%)}&
  \multicolumn{1}{l}{41.007 (-29.5\%)}&
  \multicolumn{1}{l}{11.100 (-1.2\%)}&
  \multicolumn{1}{l}{0.221 (-8.1\%)}\\
      \multicolumn{1}{l}{} &
  \multicolumn{1}{l}{Multivariate\_DAE\_2} &
  \multicolumn{1}{l}{16.553 (-20.5\%)}&
  \multicolumn{1}{l}{8.811 (-9.7\%)}&
  \multicolumn{1}{l}{0.627 (+87.9\%)}&
  \multicolumn{1}{l}{24.339 (-58.1\%)}&
  \multicolumn{1}{l}{9.508 (-15.4\%)}&
  \multicolumn{1}{l}{0.218 (-9.4\%)}\\
      \multicolumn{1}{l}{} &
  \multicolumn{1}{l}{PI-DAE} &
  \multicolumn{1}{l}{\textbf{16.436 (-21.0\%)}}&
  \multicolumn{1}{l}{8.767 (-10.1\%)}&
  \multicolumn{1}{l}{0.595 (+78.3\%)}&
  \multicolumn{1}{l}{\textbf{24.042 (-58.6\%)}}&
  \multicolumn{1}{l}{\textbf{9.362 (-16.7\%)}}&
  \multicolumn{1}{l}{0.210 (-12.4\%)}\\
   \toprule  
\end{tabular}
\end{sidewaystable}

As far as Case 1 is concerned, it is noted that the Linear Interpolation model outperforms all the considered methods on the indoor air temperature data. This behavior is significantly different for the other two variables. On the one hand, the univariate Autoencoder performs better on the heating data, reducing the RMSE by up to 16.4\% at higher training rates. On the other hand, Multivariate\_DAE\_2 and PI-DAE surpass the considered baseline with comparable performance on the cooling data, reducing the RMSE by up to 21\% at higher training rates. However, it is noted that, in general, the use of a physics-informed loss function provides only marginal improvement over Multivariate\_DAE\_2. 

Concerning Case 2, the used benchmarks outperform all the DAE's configurations at a 10\% training rate. In particular, KNN impute performs better than the baseline on both the cooling and indoor air temperature data, with a reduction of RMSE by 41.5\% and 3\%, respectively. On the other hand, the Linear Interpolation model outperforms the other methods on the heating data. It is observed that increasing the training set by just a few days can significantly increase the performance of all the Autoencoder configurations. In particular, PI-DAE surpasses all the analyzed models on the cooling flow rate, reducing the RMSE by up to 58.6\% at higher training rates. However, when considering the same model without a physics-informed loss function, namely Multivariate\_DAE\_2, the RMSE increase is barely 1\%. While the same considerations apply to the heating data, a noticeable impact of the physics-informed loss function is noticed only for the indoor air temperature data. In this regard, the RMSE decreases by up to 14.9\% and 9.7\% (30\% training rate), respectively, for PI-DAE and Multivariate\_DAE\_2. However, the univariate Autoencoder can generally perform better, reducing the RMSE by up to 26.1\% at higher training rates.

\begin{figure}[ht]
\centering
\includegraphics[width=1\textwidth]{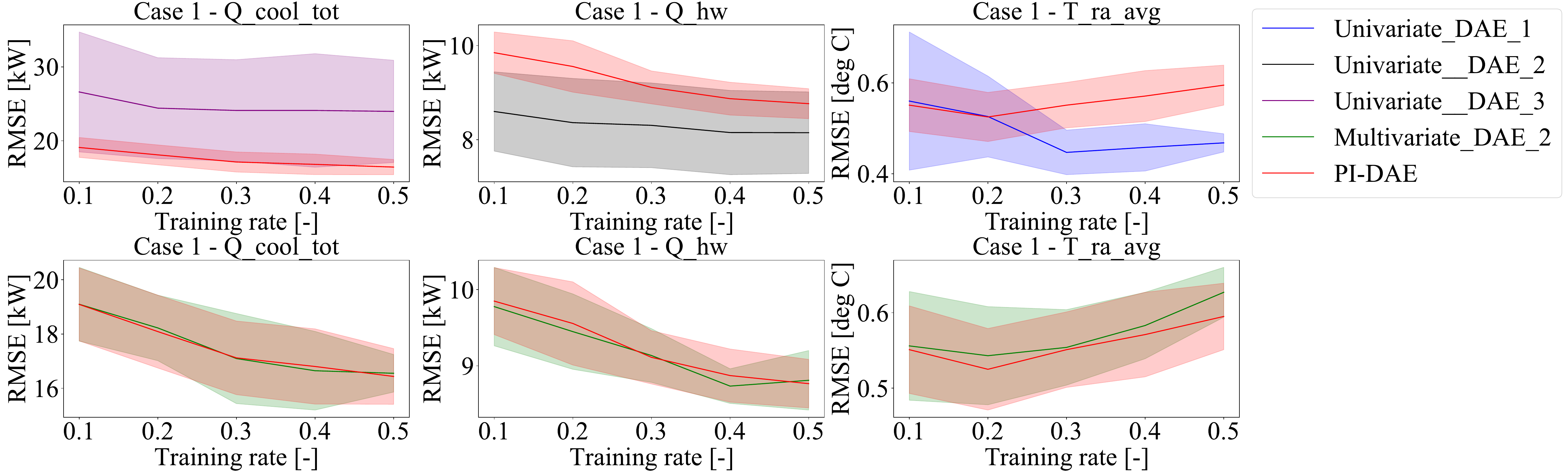}
\caption{Learning curves of the models with standard deviation for Case 1 at varied training rates. For Case 1, the depicted training rates (from 0.1 to 0.5) correspond to training set sizes of 36, 72, 108, 145, and 181 days, respectively. Alternatively, for Case 2, the depicted training rates (from 0.1 to 0.5) correspond to training set sizes of 1, 3, 5, 7, and 9 days, respectively. The analyzed variables are the total cooling flow rate ($\dot{Q}_{cool_{tot}}$), the reheat water flow rate ($\dot{Q}_{hw}$) and the average indoor air temperature ($T_{ra_{avg}}$).}
\label{fig:results_2_case1}
\end{figure}
\begin{figure}[ht]
\centering
\includegraphics[width=1\textwidth]{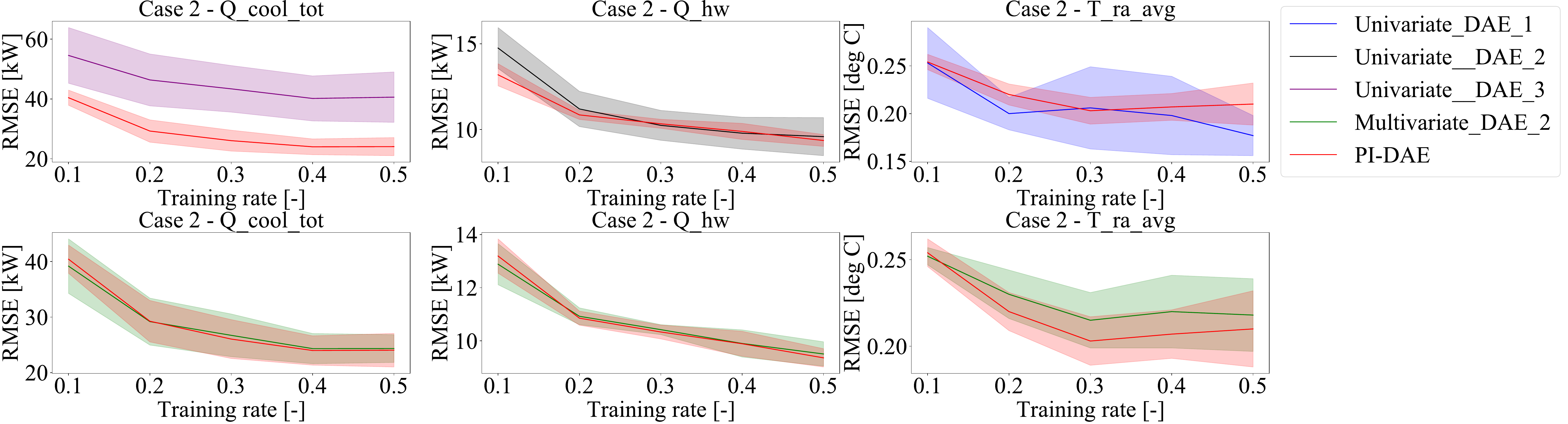}
\caption{Learning curves of the models with standard deviation for Case 2 at varied training rates and Case 2. For Case 1, the depicted training rates (from 0.1 to 0.5) correspond to training set sizes of 36, 72, 108, 145, and 181 days, respectively. Alternatively, for Case 2, the depicted training rates (from 0.1 to 0.5) correspond to training set sizes of 1, 3, 5, 7, and 9 days, respectively. The analyzed variables are the total cooling flow rate ($\dot{Q}_{cool_{tot}}$), the reheat water flow rate ($\dot{Q}_{hw}$) and the average indoor air temperature ($T_{ra_{avg}}$).}
\label{fig:results_2_case2}
\end{figure}

In order to analyze how the performance of the Autoencoders changes over different corruption rates, the standard deviation of the reconstruction errors is presented in Figures \ref{fig:results_2_case1} and \ref{fig:results_2_case2}, respectively, for Case 1 and 2. Here, Multivariate\_DAE\_1 is not represented due to the high RMSEs shown in Figure \ref{fig:results_1}. The numerical values of the reconstruction errors at different corruption rates can be found in the open-source repository$^{\ref{footnote 1}}$. It is observed that the RMSE of the univariate Autoencoder has a much higher standard deviation compared to PI-DAE. This indicates that the reconstruction error of PI-DAE is more stable along different corruption rates. However, the RMSE of PI-DAE has a standard deviation that is comparable to the same model configuration without physics-informed loss, i.e., Multivariate\_DAE\_2.

\section{Discussion}
\label{sec: discussion}

\subsection{Optimized physics-based coefficients}

\begin{figure}[ht]
\centering
\includegraphics[width=.8\textwidth]{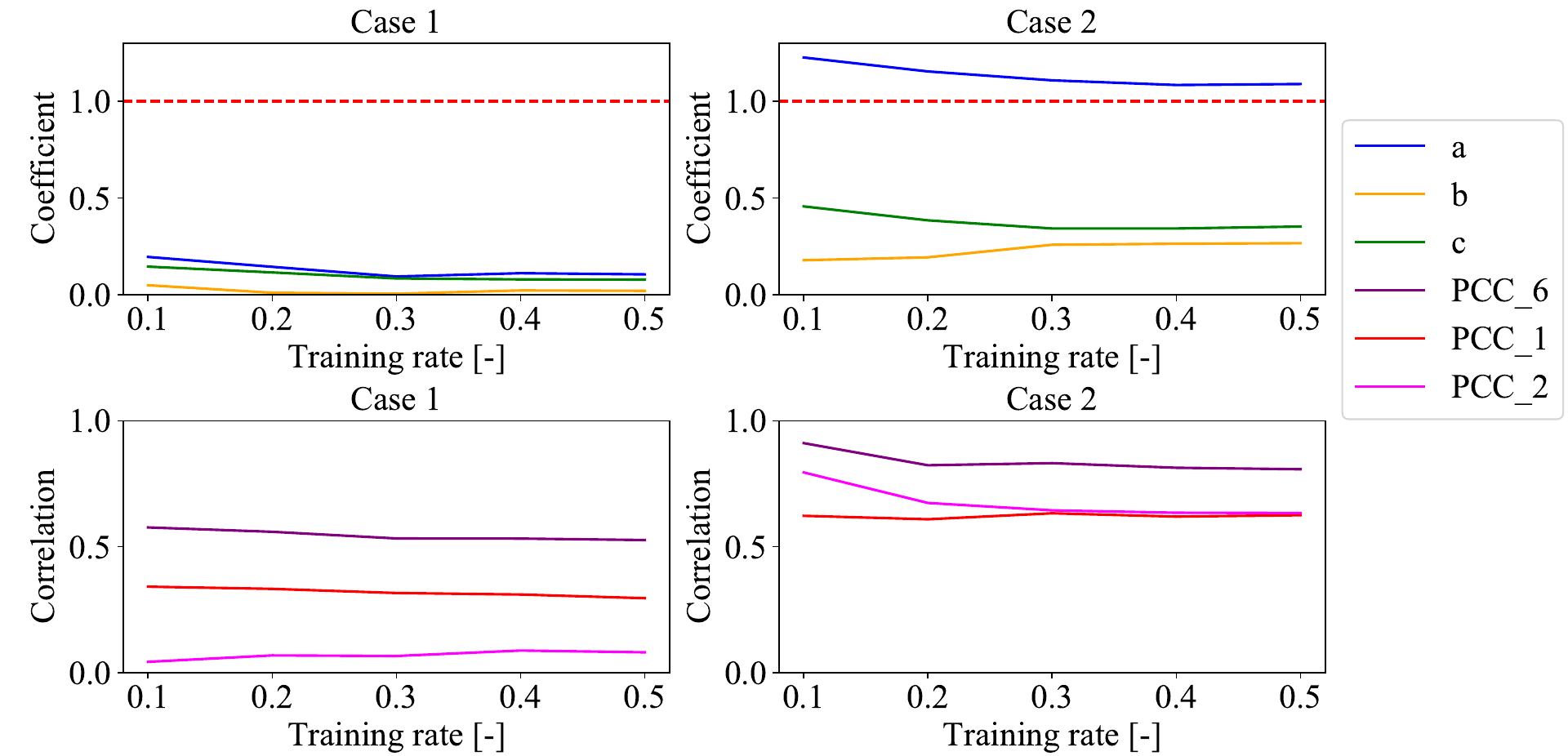}
\caption{Optimized physics-based and Pearson correlation coefficients at varied training rates and cases. For Case 1, the depicted training rates (from 0.1 to 0.5) correspond to training set sizes of 36, 72, 108, 145, and 181 days, respectively. Alternatively, for Case 2, the depicted training rates (from 0.1 to 0.5) correspond to training set sizes of 1, 3, 5, 7, and 9 days, respectively. The starting value of one for the physics-based coefficients is represented by the dashed horizontal. $PCC_6$ corresponds to the correlation coefficient between the outdoor air temperature and indoor air temperature; $PCC_1$ corresponds to the correlation coefficient between the indoor air temperature and cooling flow rate; $PCC_2$ corresponds to the correlation coefficient between the indoor air temperature and the heating flow rate.}
\label{fig:results_3}
\end{figure}

An important feature of PI-DAE is that the unknown physics-based coefficients can be optimized together with the parameters of the black-box component. In this sense, the model has the twofold objective of data imputation and system identification \cite{raissi2019physics}. For that purpose, the final coefficients are presented in Figure \ref{fig:results_3} (top part), while the numerical values can be observed in Table \ref{tab:results_3} (top part). Prior to the model's training, all the coefficients have been initialized to one as they usually tend to achieve small values after optimization \cite{gokhale2022physics,di2022physically}. It is noted that the optimized coefficients of Case 1 are much smaller than the ones of Case 2. Additionally, these are not the same for varied training rates. 

\begin{table}[ht]
\centering
\caption{Optimized physics-based and Pearson correlation coefficients at varied training rates (TR) and cases. For Case 1, the depicted training rates (from 10 to 50\%) correspond to training set sizes of 36, 72, 108, 145, and 181 days, respectively. Alternatively, for Case 2, the depicted training rates (from 10 to 50\%) correspond to training set sizes of 1, 3, 5, 7, and 9 days, respectively. $PCC_6$ corresponds to the correlation coefficient between the outdoor air temperature and indoor air temperature; $PCC_1$ corresponds to the correlation coefficient between the indoor air temperature and cooling flow rate; $PCC_2$ corresponds to the correlation coefficient between the indoor air temperature and the heating flow rate.
}
\label{tab:results_3}
\begin{tabular}{lllllll}\\
      \toprule
\multicolumn{1}{l}{TR} &
\multicolumn{3}{l}{Case 1} &
\multicolumn{3}{l}{Case 2} \\  
  \cmidrule{2-7}
  \multicolumn{1}{l}{} &
\multicolumn{1}{l}{$a$} &
  \multicolumn{1}{l}{$b$} &
  \multicolumn{1}{l}{$c$} &
  \multicolumn{1}{l}{$a$} &
  \multicolumn{1}{l}{$b$} &
  \multicolumn{1}{l}{$c$}\\
     \toprule

10\%  & 0.196 & 0.050 & 0.146 & 1.226  & 0.179 & 0.457 \\
20\%  &  0.145&  0.011&  0.116&   1.154&  0.194& 0.385  \\
30\%  &  0.095& 0.006 &  0.085&   1.108&  0.259&  0.343\\
40\%  &  0.112&  0.023& 0.080 &   1.084&  0.264& 0.343 \\
50\%  &  0.106&  0.021&  0.079&   1.089&  0.267& 0.353 \\
      \toprule
\multicolumn{1}{l}{TR} &
\multicolumn{3}{l}{Case 1} &
\multicolumn{3}{l}{Case 2} \\  
  \cmidrule{2-7}
  \multicolumn{1}{l}{} &
\multicolumn{1}{l}{$PCC_6$} &
  \multicolumn{1}{l}{$PCC_1$} &
  \multicolumn{1}{l}{$PCC_2$} &
  \multicolumn{1}{l}{$PCC_6$} &
  \multicolumn{1}{l}{$PCC_1$} &
  \multicolumn{1}{l}{$PCC_2$}\\
     \toprule

10\%&  0.576&  0.341&  -0.042&  0.911&   0.622&  -0.795 \\
20\%&  0.559&  0.332&  -0.068&  0.823&   0.609&  -0.673 \\
30\%&  0.532&  0.316&  -0.065&  0.832&   0.632&  -0.644 \\
40\%&  0.532&  0.310&  -0.087&  0.813&   0.619&  -0.635 \\
50\%&  0.526&  0.295&  -0.080&  0.807&   0.625&  -0.633 \\
      \toprule
      
\end{tabular}
\end{table}

Due to the marked variability in the estimate of the coefficients, no strict physical meaning can be assigned to them. A similar consideration was made by Brastein et al. \cite{brastein2018parameter}, although for grey-box modeling. In that case, two major causes were discussed. Namely, there is a lack of dynamic information in the training data and the presence of unmodeled disturbances in the physics-based equation. Specifically, the authors claimed that the physics-based coefficients of the grey-box model should always converge to the same values to reasonably resemble the building thermal parameters.

In this paper, the previous considerations are supported by additional findings, hence extending the concepts to physics-informed modeling. Specifically, it is noted how increasing the training rate over 30\% can considerably reduce the dispersion of the optimized coefficients, almost obtaining constant values. This observation matches intuition, as a smaller dataset would likely contain less dynamic information about the analyzed building. However, despite converging to constant values for higher training rates, the learned physics-based coefficients are still significantly different between Case 1 and 2. As discussed in Section \ref{subsec:monitoring}, it is plausible that the absence of the stochastic component in Equation \ref{eq:final_fw} forces PI-DAE to consider the unsensed effects on the indoor air temperature data, e.g., occupant behavior, by means of the varying physics-based coefficients. In this sense, smaller optimized coefficients, as in Case 1, indicate that there is a poor correlation between the temporal evolution of the indoor air temperature and the rest of the analyzed variables (see Equation \ref{eq:final_fw}). Additionally, the impact of the outdoor air temperature is much more significant than the other variables, especially for Case 2. This consideration is based on the coefficient $a$, which is always greater than the starting value.

In order to support the previous analysis, the Pearson Correlation Coefficients between the indoor air temperature and each of the other variables are depicted in Figure \ref{fig:results_3} (bottom part) and Table \ref{tab:results_3} (bottom part), at varied training rates. For better representation, coefficients with negative correlations, i.e., $PCC_2$, are plotted with positive values. It is observed that the evolution of these coefficients is very similar to the physics-based ones. In particular, the dispersion of the $PCCs$ tends to reduce for higher training rates, while a remarkable difference is noted between Case 1 and 2. However, the two behaviors are not exactly the same. For instance, the results from PI-DAE suggest that the cooling flow rate, i.e., $b$, has the least influence over the indoor air temperature data for Case 1. Alternatively, the Pearson Correlation Coefficient between the indoor air temperature and the cooling flow rate, i.e., $PCC_1$, has the second major influence. This discrepancy might be explained by assuming that PI-DAE identifies variables that are correlated not only statistically, as the Pearson Correlation Coefficients, but also physically through Equation \ref{eq:final_fw}.

\begin{figure}[ht]
\centering
\includegraphics[width=.9\textwidth]{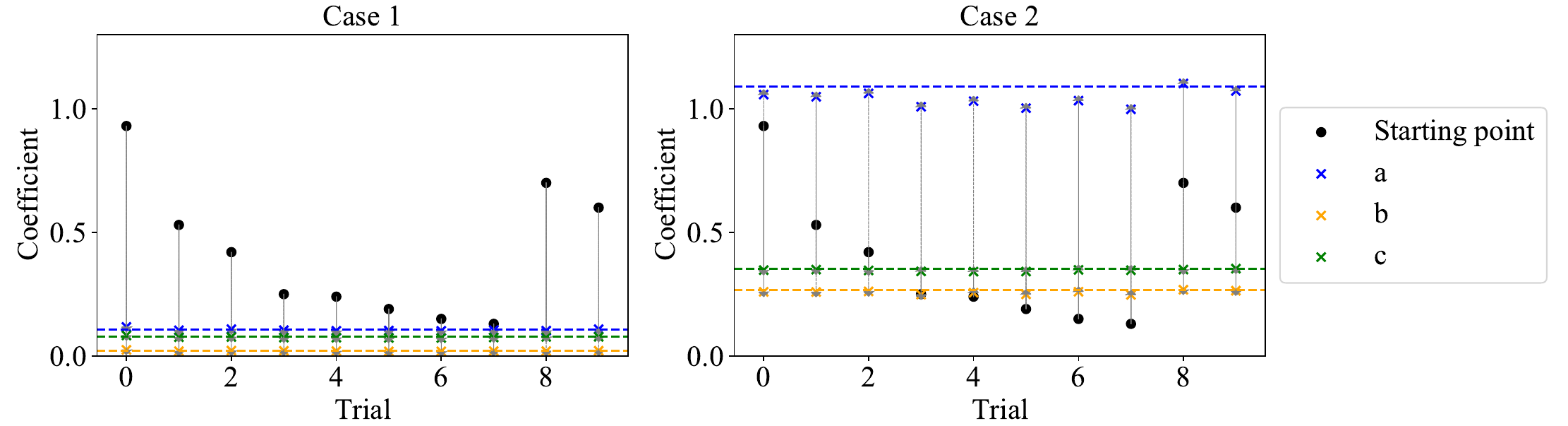}
\caption{Optimized physics-based coefficients under different starting points for a training rate of 0.5. The starting points are represented by the black dots. The x-markers represent the final physics-based coefficients after optimization. The dashed horizontal lines are the final optimized coefficients of Table \ref{tab:results_3}.}
\label{fig:results_6}
\end{figure}

Based on the previous considerations, it is observed that the analyzed physics-based coefficients should always converge to constant values, provided that the training data contain enough dynamic information (i.e., the dataset is big enough) and the same amount of unsensed disturbances (i.e., same building monitoring period). As confirmed by Brastein et al. \cite{brastein2018parameter}, this should hold for any initial guess of these coefficients. For that purpose, Figure \ref{fig:results_6} depicts the optimized physics-based coefficients under different starting points for a training rate of 0.5. As explained earlier, these coefficients usually tend to achieve small values. Therefore, the starting points should be small enough to avoid any convergence issues during training. For this reason, these are randomly selected ten times between zero and one by using the related ``random.random" function from the numpy library. The results indicate a really low degree of dispersion, hence further proving the enhanced interpretability of PI-DAE. In particular, the highest variability is noted for the coefficient $a$ of Case 2. Here, the difference is more pronounced for trials whose starting points are further from the optimal value. This can be explained by considering that models using optimization methods based on the stochastic gradient descent, such as PI-DAE, make repeated small incremental updates to the training parameters during the optimization process \cite{goodfellow2016deep}. Therefore, the physics-based coefficients are likely to not converge if starting from points far from the optimal values.

In summary, PI-DAE is capable of identifying physically consistent multi-dimensional correlations among the input features. The novelty of this finding lies in the relationship between the optimized physics-based coefficients and the final reconstruction errors, hence enhancing the inherent model interpretability. Namely, it is shown that smaller physics-based coefficients indicate a greater presence of disturbances external to the observed environment. On the other hand, greater physics-based coefficients indicate a smaller presence of disturbances external to the observed environment. This greatly affects the model performance when applying a multivariate configuration, thus when exploiting multi-dimensional correlations among the considered variables (see Section \ref{sec:results}). However, the analyzed dataset should contain enough dynamic information to ensure consistent convergence, i.e., without dispersion, of the physics-based coefficients.

\subsection{Computational requirements}

\begin{figure}[ht]
\centering
\includegraphics[width=1\textwidth]{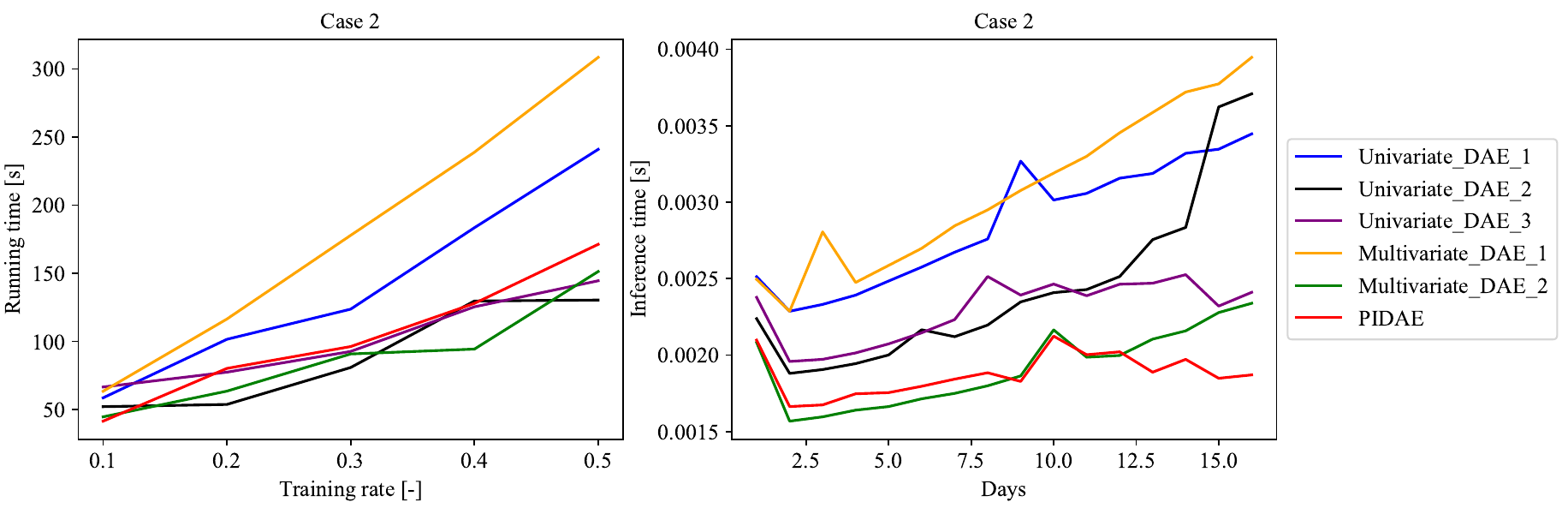}
\caption{Running and inference times of the models respectively at varied training rates and days for Case 2. The depicted training rates (from 0.1 to 0.5) correspond to training set sizes of 1, 3, 5, 7, and 9 days, respectively.}
\label{fig:results_3_1}
\end{figure}

In order to quantify the impact of the physics-informed loss function on the computational requirements of the multivariate Autoencoder, the running and inference times of the analyzed DAE configurations can be observed in Figure \ref{fig:results_3_1}. Since a comparison between Case 1 and Case 2 is irrelevant, i.e., Case 1 would be more computationally intensive due to the larger dataset, only Case 2 is selected. Here, the running time is the time required by the models prior to being exported for evaluation. As explained in Section \ref{sec:design}, this corresponds to the total time needed to train and validate the models ten times. On the other hand, the inference time is the time required by the models to impute an increasing number of days on the evaluation set \cite{liguori2023augmenting}. For Case 2, the total number of days on the evaluation set varies from 8 (50\% training rate) to 16 (10\% training rate). For this reason, the inference time is averaged for the first days, while the last days are only related to the models with the 10\% training rate.

It is noted that the level of complexity of the optimized models, i.e., the number of trainable parameters, has a significant impact on both the running and inference times. This finding matches intuition, as an excessively complex model is expected to experience higher computational and time costs compared to more simplified counterparts \cite{hu2021model}. Specifically, Univariate\_DAE\_1 and Multivariate\_DAE\_1 have both the greatest number of trainable parameters and required computational resources. Even if there is not a direct correlation between the two factors, this analysis proves that the computational complexity introduced by the physical component of PI-DAE is not as relevant as the one derived from the hyperparameter tuning of Section \ref{subsec: development}.

\subsection{Prospects and limitations}

The analysis of Section \ref{sec:results} confirmed the importance of selecting appropriate building operation periods before applying PI-DAE. Namely, the temporal profiles of the features that are used should be highly correlated with each other to reduce the effects of the unmodeled disturbances. However, this consideration also applies to the same model configuration without a physical component, namely Multivariate\_DAE\_2. In this sense, the use of a physics-informed loss function provides only marginal performance improvement. It is plausible that Case 2 still contains a significant number of unsensed effects that impact data imputation. Accordingly, it is noted how the univariate models can, on the other hand, generally perform better or comparably well than PI-DAE or Multivariate\_DAE\_2. The only exception is for the cooling data, for which correlations with the other variables are higher. This supports the consideration that multivariate methods are strongly affected by the presence of unseen disturbances since they tend to rely heavily on the multi-dimensional correlation attributes of the input features \cite{loy2020imputing}. For instance, the lower reconstruction error experienced by the univariate model on the indoor air temperature indicates that, for this variable, the temporal correlations are more important than the spatial ones.

When the training set is as little as one day, simplified imputation approaches, such as the Linear Interpolation or KNN Impute, can generally provide more accurate replacements for missing data. The same consideration can be made for low variability data, such as the indoor air temperature profiles of Case 1. On the other hand, when increasing the number of training days, convergence issues might occur on the indoor air temperature data, especially for PI-DAE and Multivariate\_DAE\_2. Although this behavior may seem counter-intuitive, it is again attributable to the imperfect correlation between the analyzed variables. As a matter of fact, increasing the number of training data might introduce correlations that are present at the time of training but not at the time of evaluation. A distribution shift phenomenon which is generally referred to as spurious correlation \cite{wiles2021fine}.

In summary, it is essential for the variables used to be highly correlated with each other in order to apply PI-DAE effectively. In the present work, it is proven 
that the optimized physics-based coefficients of Equation \ref{eq:final_fw} can indeed provide insight into the extent to which the presence of unmodeled disturbances can impact the reconstruction error of the model. Although the univariate Autoencoders are, in many cases, better performing than the proposed approach, they lack this enhanced inherent interpretability. However, additional studies are needed in this direction as the model cannot yet be defined as fully inherently interpretable. For instance, a significant downside of this paper is that it performs a relative rather than an absolute analysis. That is, for a specific building monitoring dataset, it is only possible to quantify which period might impact more the imputation process, i.e. which period experiences the greatest amount of external disturbances. In this sense, if the correlation among the analyzed features is not dynamic, i.e., no significant variation of the unsensed effects can be noted in the dataset, the optimized physics-based coefficients will not provide useful information to the users. Therefore, it would be much more useful to define a threshold above which the performance of the model starts to decrease. However, this threshold would logically depend on the specific building's thermal properties.

\subsection{Practical implications}

The application of the proposed PI-DAE in missing data imputation for commercial buildings has practical implications for professionals in the building industry. This approach effectively addresses the challenges of missing data and limited training datasets commonly encountered in building systems. By incorporating physics-inspired constraints into the loss function of the Denoising Autoencoder, the PI-DAE model enables stable and effective data imputation, even when working with relatively limited quantities of data. This is particularly valuable in the building industry, where data is often incomplete, and training datasets may be insufficient. Above all, the optimized physics-based coefficients provide important physical insights into the analyzed building monitoring dataset. Specifically, they help to quantify the presence of unmodeled disturbances and their impact on the data imputation process. This is of utmost importance in commercial buildings, where both accurate and reliable data are essential for decision-making related to energy management, retrofit analysis, and overall optimization of building performance. It is argued that this enhanced inherent model interpretability might represent a significant step forward toward bridging the gap between academic research and industry.

Moreover, this approach can extend beyond the targeted variables of heating, cooling, and indoor air temperature, which helps downstream evaluation of the effectiveness of retrofit measures and maintenance of indoor comfort. The proposed method can be further applied to other time series data from building systems, such as HVAC, lighting, and appliances data, as long as they also adhere to underlying physical principles. This opens up possibilities for applying the proposed method on a larger scale. By leveraging the power of physics-informed constraints, the PI-DAE model holds the potential to address data imputation challenges within the building industry.

\section{Conclusion}
\label{sec: conclusion}

This paper presented a study to quantify the effects of embedding physical knowledge in a Denoising Autoencoder for missing data imputation. The proposed PI-DAE combines the potential of DAEs with an approximated building thermal balance ODE, which enforces physical laws on the reconstructed features. By guiding the imputed indoor air temperature, heating, and cooling data within physically meaningful boundaries, the presented model could support a building retrofit analysis. For that purpose, it is observed that the unknown physics-based coefficients can be optimized together with the parameters of the black-box component. The relevant findings of this study may be summarized as follows:


\begin{itemize}

\item The RMSE of PI-DAE could be reduced by up to 58.6\% (cooling data), 16.7\% (heating data), and 14.9\% (indoor air temperature data) with respect to the selected baseline.
\item The imputation robustness of PI-DAE could be significantly improved over different missing intervals with respect to more simplified DAE configurations.
\item The optimized physics-based coefficients converged to constant values on different building monitoring periods.
\item The computational complexity introduced by the physical module was much less significant than the one introduced by the black box component due to the integration of physics-informed constraints.

\end{itemize}

In summary, the results revealed that the use of a physics-informed loss function provides only marginal improvements compared to the base model configuration. However, the optimized physics-based coefficients provide important insights into the building monitoring periods characterized by unmodeled disturbances. Here, higher coefficients correspond to intervals of time when the used variables are highly correlated with each other. This integration notably enhances the robustness of training, especially when working with a limited dataset. Additionally, the discovered optimized physics-based coefficients play a vital role in shedding light on the underlying physics of the black-box models for the imputed variables.

\section{Acknowledgements}This work was funded by the Deutsche Forschungsgemeinschaft (DFG, German Research Foundation) – TR 892/8-1. Simulations were performed with computing resources granted by RWTH Aachen University, Germany, under project rwth0622.

\bibliographystyle{elsarticle-num}
\bibliography{main}

\begin{thebibliography}{10}
\expandafter\ifx\csname url\endcsname\relax
  \def\url#1{\texttt{#1}}\fi
\expandafter\ifx\csname urlprefix\endcsname\relax\def\urlprefix{URL }\fi
\expandafter\ifx\csname href\endcsname\relax
  \def\href#1#2{#2} \def\path#1{#1}\fi

\bibitem{fetting2020european}
C.~Fetting, The european green deal, ESDN Report, December (2020).

\bibitem{thrampoulidis2023approximating}
E.~Thrampoulidis, G.~Hug, K.~Orehounig, Approximating optimal building retrofit
  solutions for large-scale retrofit analysis, Applied Energy 333 (2023)
  120566.

\bibitem{chong2021calibrating}
A.~Chong, Y.~Gu, H.~Jia, Calibrating building energy simulation models: A
  review of the basics to guide future work, Energy and Buildings 253 (2021)
  111533.

\bibitem{angelotti2018building}
A.~Angelotti, M.~Martire, L.~Mazzarella, M.~Pasini, I.~Ballarini, V.~Corrado,
  G.~De~Luca, P.~Baggio, A.~Prada, F.~Bosco, et~al., Building energy simulation
  for nearly zero energy retrofit design: the model calibration, in: 2018 IEEE
  International Conference on Environment and Electrical Engineering and 2018
  IEEE Industrial and Commercial Power Systems Europe (EEEIC/I\&CPS Europe),
  IEEE, 2018, pp. 1--6.

\bibitem{magnier2010multiobjective}
L.~Magnier, F.~Haghighat, Multiobjective optimization of building design using
  trnsys simulations, genetic algorithm, and artificial neural network,
  Building and Environment 45~(3) (2010) 739--746.

\bibitem{ji2015bottom}
Y.~Ji, P.~Xu, A bottom-up and procedural calibration method for building energy
  simulation models based on hourly electricity submetering data, Energy 93
  (2015) 2337--2350.

\bibitem{claridge2006missing}
D.~E. Claridge, H.~Chen, Missing data estimation for 1--6 h gaps in energy use
  and weather data using different statistical methods, International journal
  of energy research 30~(13) (2006) 1075--1091.

\bibitem{baltazar2002restoration}
J.~C. Baltazar, D.~E. Claridge, Restoration of short periods of missing energy
  use and weather data using cubic spline and fourier series approaches:
  qualitative analysis (2002).

\bibitem{liguori2021indoor}
A.~Liguori, R.~Markovic, T.~T.~H. Dam, J.~Frisch, C.~van Treeck, F.~Causone,
  Indoor environment data time-series reconstruction using autoencoder neural
  networks, Building and Environment 191 (2021) 107623.

\bibitem{donders2006gentle}
A.~R.~T. Donders, G.~J. Van Der~Heijden, T.~Stijnen, K.~G. Moons, A gentle
  introduction to imputation of missing values, Journal of clinical
  epidemiology 59~(10) (2006) 1087--1091.

\bibitem{emmanuel2021survey}
T.~Emmanuel, T.~Maupong, D.~Mpoeleng, T.~Semong, B.~Mphago, O.~Tabona, A survey
  on missing data in machine learning, Journal of Big Data 8~(1) (2021) 1--37.

\bibitem{liu2015fault}
Z.~Liu, Y.~Liu, D.~Zhang, B.~Cai, C.~Zheng, Fault diagnosis for a solar
  assisted heat pump system under incomplete data and expert knowledge, Energy
  87 (2015) 41--48.

\bibitem{li2020missing}
H.~Li, X.~Chen, M.~Shan, P.~Duan, Missing data filling methods of
  air-conditioning power consumption for public buildings, in: 2020 39th
  Chinese Control Conference (CCC), IEEE, 2020, pp. 3183--3187.

\bibitem{wang2021fault}
Z.~Wang, L.~Wang, Y.~Tan, J.~Yuan, Fault detection based on bayesian network
  and missing data imputation for building energy systems, Applied Thermal
  Engineering 182 (2021) 116051.

\bibitem{fu2024filling}
C.~Fu, M.~Quintana, Z.~Nagy, C.~Miller, Filling time-series gaps using image
  techniques: Multidimensional context autoencoder approach for building energy
  data imputation, Applied Thermal Engineering 236 (2024) 121545.

\bibitem{ahn2022comparison}
H.~Ahn, K.~Sun, K.~P. Kim, Comparison of missing data imputation methods in
  time series forecasting, Computers, Materials \& Continua 70~(1) (2022)
  767--779.

\bibitem{hussain2022novel}
S.~N. Hussain, A.~A. Aziz, M.~J. Hossen, N.~A.~A. Aziz, G.~R. Murthy, F.~B.
  Mustakim, A novel framework based on cnn-lstm neural network for prediction
  of missing values in electricity consumption time-series datasets, Journal of
  Information Processing Systems 18~(1) (2022) 115--129.

\bibitem{festag2022generative}
S.~Festag, J.~Denzler, C.~Spreckelsen, Generative adversarial networks for
  biomedical time series forecasting and imputation, Journal of Biomedical
  Informatics 129 (2022) 104058.

\bibitem{ma2020bi}
J.~Ma, J.~C. Cheng, F.~Jiang, W.~Chen, M.~Wang, C.~Zhai, A bi-directional
  missing data imputation scheme based on lstm and transfer learning for
  building energy data, Energy and Buildings 216 (2020) 109941.

\bibitem{arjunan2020energystar++}
P.~Arjunan, K.~Poolla, C.~Miller, Energystar++: Towards more accurate and
  explanatory building energy benchmarking, Applied Energy 276 (2020) 115413.

\bibitem{yao2013state}
Y.~Yao, K.~Yang, M.~Huang, L.~Wang, A state-space model for dynamic response of
  indoor air temperature and humidity, Building and environment 64 (2013)
  26--37.

\bibitem{karniadakis2021physics}
G.~E. Karniadakis, I.~G. Kevrekidis, L.~Lu, P.~Perdikaris, S.~Wang, L.~Yang,
  Physics-informed machine learning, Nature Reviews Physics 3~(6) (2021)
  422--440.

\bibitem{raissi2019physics}
M.~Raissi, P.~Perdikaris, G.~E. Karniadakis, Physics-informed neural networks:
  A deep learning framework for solving forward and inverse problems involving
  nonlinear partial differential equations, Journal of Computational physics
  378 (2019) 686--707.

\bibitem{gokhale2022physics}
G.~Gokhale, B.~Claessens, C.~Develder, Physics informed neural networks for
  control oriented thermal modeling of buildings, Applied Energy 314 (2022)
  118852.

\bibitem{di2022physically}
L.~Di~Natale, B.~Svetozarevic, P.~Heer, C.~N. Jones, Physically consistent
  neural networks for building thermal modeling: theory and analysis, Applied
  Energy 325 (2022) 119806.

\bibitem{di2023towards}
L.~Di~Natale, B.~Svetozarevic, P.~Heer, C.~N. Jones, Towards scalable
  physically consistent neural networks: An application to data-driven
  multi-zone thermal building models, Applied Energy 340 (2023) 121071.

\bibitem{xiao2023building}
T.~Xiao, F.~You, Building thermal modeling and model predictive control with
  physically consistent deep learning for decarbonization and energy
  optimization, Applied Energy 342 (2023) 121165.

\bibitem{nagarathinam2022pacman}
S.~Nagarathinam, Y.~S. Chati, M.~P. Venkat, A.~Vasan, Pacman: physics-aware
  control manager for hvac, in: Proceedings of the 9th ACM International
  Conference on Systems for Energy-Efficient Buildings, Cities, and
  Transportation, 2022, pp. 11--20.

\bibitem{wang2023physics}
X.~Wang, B.~Dong, Physics-informed hierarchical data-driven predictive control
  for building hvac systems to achieve energy and health nexus, Energy and
  Buildings (2023) 113088.

\bibitem{chen2023physics}
Y.~Chen, Q.~Yang, Z.~Chen, C.~Yan, S.~Zeng, M.~Dai, Physics-informed neural
  networks for building thermal modeling and demand response control, Building
  and Environment 234 (2023) 110149.

\bibitem{caruana2015intelligible}
R.~Caruana, Y.~Lou, J.~Gehrke, P.~Koch, M.~Sturm, N.~Elhadad, Intelligible
  models for healthcare: Predicting pneumonia risk and hospital 30-day
  readmission, in: Proceedings of the 21th ACM SIGKDD international conference
  on knowledge discovery and data mining, 2015, pp. 1721--1730.

\bibitem{bishop2006pattern}
C.~M. Bishop, N.~M. Nasrabadi, Pattern recognition and machine learning,
  Vol.~4, Springer, 2006.

\bibitem{fan2019novel}
C.~Fan, F.~Xiao, C.~Yan, C.~Liu, Z.~Li, J.~Wang, A novel methodology to explain
  and evaluate data-driven building energy performance models based on
  interpretable machine learning, Applied Energy 235 (2019) 1551--1560.

\bibitem{deb2021review}
C.~Deb, A.~Schlueter, Review of data-driven energy modelling techniques for
  building retrofit, Renewable and Sustainable Energy Reviews 144 (2021)
  110990.

\bibitem{chen2023interpretable}
Z.~Chen, F.~Xiao, F.~Guo, J.~Yan, Interpretable machine learning for building
  energy management: A state-of-the-art review, Advances in Applied Energy
  (2023) 100123.

\bibitem{chen2023missing}
Z.~Chen, S.~Tan, U.~Chajewska, C.~Rudin, R.~Caruna, Missing values and
  imputation in healthcare data: Can interpretable machine learning help?, in:
  Conference on Health, Inference, and Learning, PMLR, 2023, pp. 86--99.

\bibitem{brumley2006towards}
D.~Brumley, D.~Song, J.~Slember, Towards automatically eliminating
  integer-based vulnerabilities (2006).

\bibitem{chong2016imputation}
A.~Chong, K.~P. Lam, W.~Xu, O.~T. Karaguzel, Y.~Mo, Imputation of missing
  values in building sensor data, ASHRAE and IBPSA-USA SimBuild 6 (2016)
  407--14.

\bibitem{mishra2017local}
S.~Mishra, B.~L. Sturm, S.~Dixon, Local interpretable model-agnostic
  explanations for music content analysis., in: ISMIR, Vol.~53, 2017, pp.
  537--543.

\bibitem{sudjianto2021designing}
A.~Sudjianto, A.~Zhang, Designing inherently interpretable machine learning
  models, arXiv preprint arXiv:2111.01743 (2021).

\bibitem{li2019novel}
D.~Li, D.~Li, C.~Li, L.~Li, L.~Gao, A novel data-temporal attention network
  based strategy for fault diagnosis of chiller sensors, Energy and Buildings
  198 (2019) 377--394.

\bibitem{liguori2023augmenting}
A.~Liguori, R.~Markovic, M.~Ferrando, J.~Frisch, F.~Causone, C.~van Treeck,
  Augmenting energy time-series for data-efficient imputation of missing
  values, Applied Energy 334 (2023) 120701.

\bibitem{luo2022three}
N.~Luo, Z.~Wang, D.~Blum, C.~Weyandt, N.~Bourassa, M.~A. Piette, T.~Hong, A
  three-year dataset supporting research on building energy management and
  occupancy analytics, Scientific Data 9~(1) (2022) 156.

\bibitem{wang2017physics}
J.-X. Wang, J.-L. Wu, H.~Xiao, Physics-informed machine learning approach for
  reconstructing reynolds stress modeling discrepancies based on dns data,
  Physical Review Fluids 2~(3) (2017) 034603.

\bibitem{wu2018physics}
J.-L. Wu, H.~Xiao, E.~Paterson, Physics-informed machine learning approach for
  augmenting turbulence models: A comprehensive framework, Physical Review
  Fluids 3~(7) (2018) 074602.

\bibitem{howland2019wind}
M.~F. Howland, J.~O. Dabiri, Wind farm modeling with interpretable
  physics-informed machine learning, Energies 12~(14) (2019) 2716.

\bibitem{karimpouli2020physics}
S.~Karimpouli, P.~Tahmasebi, Physics informed machine learning: Seismic wave
  equation, Geoscience Frontiers 11~(6) (2020) 1993--2001.

\bibitem{raissi2017physics}
M.~Raissi, P.~Perdikaris, G.~E. Karniadakis, Physics informed deep learning
  (part i): Data-driven solutions of nonlinear partial differential equations,
  arXiv preprint arXiv:1711.10561 (2017).

\bibitem{zobeiry2021physics}
N.~Zobeiry, K.~D. Humfeld, A physics-informed machine learning approach for
  solving heat transfer equation in advanced manufacturing and engineering
  applications, Engineering Applications of Artificial Intelligence 101 (2021)
  104232.

\bibitem{drgona2020physics}
J.~Drgona, A.~R. Tuor, V.~Chandan, D.~L. Vrabie, Physics-constrained deep
  recurrent neural models of building thermal dynamics, Tech. rep., Pacific
  Northwest National Lab.(PNNL), Richland, WA (United States) (2020).

\bibitem{blum2022field}
D.~Blum, Z.~Wang, C.~Weyandt, D.~Kim, M.~Wetter, T.~Hong, M.~A. Piette, Field
  demonstration and implementation analysis of model predictive control in an
  office hvac system, Applied Energy 318 (2022) 119104.

\bibitem{bertagnolio2008simulation}
S.~Bertagnolio, J.~Lebrun, Simulation of a building and its hvac system with an
  equation solver: application to benchmarking, in: Building Simulation,
  Vol.~1, Springer, 2008, pp. 234--250.

\bibitem{ferrari2015building}
S.~Ferrari, V.~Zanotto, Building energy performance assessment in Southern
  Europe, Springer, 2015, Ch. 1.1.

\bibitem{van2010introduction}
C.~A. van Treeck, Introduction to building performance modeling and simulation,
  Ph.D. thesis, Habilitationsschrift, Technische Universit{\"a}t M{\"u}nchen,
  2010 (2010).

\bibitem{goodfellow2016deep}
I.~Goodfellow, Y.~Bengio, A.~Courville, Deep learning, MIT press, 2016.

\bibitem{liguori2021gap}
A.~Liguori, R.~Markovic, J.~Frisch, A.~Wagner, F.~Causone, C.~van Treeck,
  et~al., A gap-filling method for room temperature data based on autoencoder
  neural networks, in: BUILDING SIMULATION CONFERENCE PROCEEDINGS, Vol.~17,
  2021, pp. 2427--2434.

\bibitem{jagtap2020adaptive}
A.~D. Jagtap, K.~Kawaguchi, G.~E. Karniadakis, Adaptive activation functions
  accelerate convergence in deep and physics-informed neural networks, Journal
  of Computational Physics 404 (2020) 109136.

\bibitem{akiba2019optuna}
T.~Akiba, S.~Sano, T.~Yanase, T.~Ohta, M.~Koyama, Optuna: A next-generation
  hyperparameter optimization framework, in: Proceedings of the 25th ACM SIGKDD
  international conference on knowledge discovery \& data mining, 2019, pp.
  2623--2631.

\bibitem{andersen2000modelling}
K.~K. Andersen, H.~Madsen, L.~H. Hansen, Modelling the heat dynamics of a
  building using stochastic differential equations, Energy and Buildings 31~(1)
  (2000) 13--24.

\bibitem{brastein2018parameter}
O.~M. Brastein, D.~W.~U. Perera, C.~Pfeifer, N.-O. Skeie, Parameter estimation
  for grey-box models of building thermal behaviour, Energy and Buildings 169
  (2018) 58--68.

\bibitem{hung2017interpretation}
M.~Hung, J.~Bounsanga, M.~W. Voss, Interpretation of correlations in clinical
  research, Postgraduate medicine 129~(8) (2017) 902--906.

\bibitem{qureshi2017wind}
A.~S. Qureshi, A.~Khan, A.~Zameer, A.~Usman, Wind power prediction using deep
  neural network based meta regression and transfer learning, Applied Soft
  Computing 58 (2017) 742--755.

\bibitem{ahmad2012efficient}
S.~Ahmad, Z.~Lin, S.~A. Abbasi, M.~Riaz, et~al., On efficient monitoring of
  process dispersion using interquartile range, Open journal of applied
  sciences 2~(04) (2012) 39--43.

\bibitem{hu2021model}
X.~Hu, L.~Chu, J.~Pei, W.~Liu, J.~Bian, Model complexity of deep learning: A
  survey, Knowledge and Information Systems 63 (2021) 2585--2619.

\bibitem{loy2020imputing}
J.~Loy-Benitez, S.~Heo, C.~Yoo, Imputing missing indoor air quality data via
  variational convolutional autoencoders: Implications for ventilation
  management of subway metro systems, Building and Environment 182 (2020)
  107135.

\bibitem{wiles2021fine}
O.~Wiles, S.~Gowal, F.~Stimberg, S.~Alvise-Rebuffi, I.~Ktena, K.~Dvijotham,
  T.~Cemgil, A fine-grained analysis on distribution shift, arXiv preprint
  arXiv:2110.11328 (2021).

\end{thebibliography}
\biboptions{sort&compress}

\end{document}